\documentclass[journal]{IEEEtran}
\usepackage{comment}
\usepackage{times}
\usepackage{epsfig}
\usepackage{graphicx}
\usepackage{amsmath}
\usepackage{amssymb}
\usepackage{mathtools}
\usepackage[export]{adjustbox}
\usepackage{subfig}
\usepackage{pifont}
\usepackage{enumitem}
\usepackage[ruled,vlined]{algorithm2e}
\usepackage{amsthm}
\usepackage{multirow}
\usepackage[table,xcdraw]{xcolor}
\usepackage[breaklinks,colorlinks]{hyperref}
\usepackage{soul}
\usepackage[noadjust]{cite}
\usepackage{alphalph}

\begin{document}
\title{Self-supervision via Controlled Transformation \\and Unpaired Self-conditioning for \\Low-light Image Enhancement}

\author{Aupendu~Kar,
        Sobhan~Kanti~Dhara,
        Debashis~Sen,~\IEEEmembership{Senior~Member,~IEEE,}
        and~Prabir~Kumar~Biswas,~\IEEEmembership{Senior~Member,~IEEE}}


\maketitle
\begin{abstract}
Real-world low-light images captured by imaging devices suffer from poor visibility and require a domain-specific enhancement to produce artifact-free outputs that reveal details. However, it is usually challenging to create large-scale paired real-world low-light image datasets for training enhancement approaches. When trained with limited data, most supervised approaches do not perform well in generalizing to a wide variety of real-world images. In this paper, we propose an unpaired low-light image enhancement network leveraging novel controlled transformation-based self-supervision and unpaired self-conditioning strategies. The model determines the required degrees of enhancement at the input image pixels, which are learned from the unpaired low-lit and well-lit images without any direct supervision. The self-supervision is based on a controlled transformation of the input image and subsequent maintenance of its enhancement in spite of the transformation. The self-conditioning performs training of the model on unpaired images such that it does not enhance an already-enhanced image or a well-lit input image. The inherent noise in the input low-light images is handled by employing low gradient magnitude suppression in a detail-preserving manner. In addition, our noise handling is self-conditioned by preventing the denoising of noise-free well-lit images. The training based on low-light image enhancement-specific attributes allows our model to avoid paired supervision without compromising significantly in performance. While our proposed self-supervision aids consistent enhancement, our novel self-conditioning facilitates adequate enhancement. Extensive experiments on multiple standard datasets demonstrate that our model, in general, outperforms the state-of-the-art both quantitatively and subjectively. Ablation studies show the effectiveness of our self-supervision and self-conditioning strategies, and the related loss functions.
\end{abstract}

\IEEEpeerreviewmaketitle


\section{Introduction}
\label{sec:intro}
Low-light images are usually captured due to low ambient light or inappropriate settings in imaging devices, and such images suffer from low scene content visibility~\cite{10146332,liu2021benchmarking, zhang2021beyond, 9609683, 9953166}. Further, due to the low-lighting conditions, the images captured by those devices are often embedded with intrinsic noise~\cite{li2018structure, liu2021benchmarking, zhang2021beyond, 9750136}. Therefore, a low-light image enhancement approach targeted to improve the content visibility, which also handles the low-light specific noise, is required for human viewing and subsequent computer vision tasks for measurements and analysis~\cite {10146332, 9609683, 9953166}.

In the last decade, there have been several proposals on generic image contrast enhancement, which in general neither perform well in low-light specific enhancement nor are they meant to do so. A plethora of investigations on enhancing low-light images exist in the literature. Most such recent approaches are learning-based~\cite{wei2018deep,liu2021retinex,yang2021sparse, yang2021band, guo2020zero,yang2020fidelity,zheng2021adaptive,jiang2021enlightengan,lee2020unsupervised, zhang2020self, xiong2020unsupervised, chen2018learning,      ren2019low} and other approaches developed in the last few years that do not use learning algorithms are based on the decomposition of images into reflectance and illumination maps through regularized optimization~\cite{Fu2016, Guo2017,   li2018structure, ren2020lr3m, wang2019low, dhara2021exposedness, Park2017}. Several such decomposition-based techniques apply predefined fixed transformations~\cite{Guo2017, li2018structure, ren2020lr3m} on the estimated image illumination, making them prone to over- and under-enhancement~\cite{liu2021retinex}. Owing to the unique noise-ingrained nature of low-light images, a majority of the deep learning based enhancement methods rely on paired training data~\cite{zheng2021adaptive, liu2021retinex, 9609683}. However, the availability of a large number of low-light and well-lit image pairs of various kinds is a challenging requirement, without which the performance may suffer, especially in real-world images~\cite{zheng2021adaptive, liu2021retinex}. 

To overcome the above issue, recently, a few deep learning based approaches have been proposed that do not require ground truth data~\cite{9609683}, among which some work with unpaired images~\cite{jiang2021enlightengan} and others only with low-light images~\cite{kar2021zero, guo2020zero,zhao2021retinexdip,liu2021retinex}. As discussed later with details in Section~\ref{sec:algo}, while most of these approaches use generic enhancement attributes and optimization losses, some of them employ assumptions about image characteristics. Further, low-light relevant aspects are ignored in noise-handling by many state-of-the-art enhancement approaches that employ generic denoising regularizers~\cite{liu2021retinex, zheng2021adaptive}. A detailed description of the recent related work is given in Section~\ref{sec:literature}.

\begin{figure}
   \centering
   \subfloat[ Unpaired Training Model for Enhancement (trained first)]{
     \includegraphics[width=0.21\textwidth]{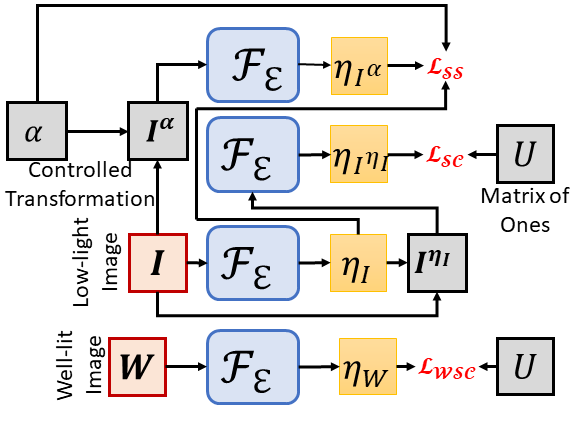}}
   \hfill
   \subfloat[ Unpaired Training Model for Noise-handling (trained last)]{
     \includegraphics[width=0.20\textwidth]{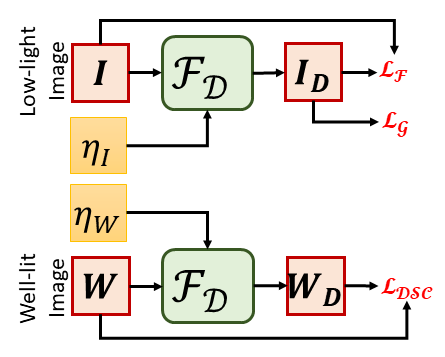}}\\
    \subfloat[ Testing Model {\color{black}(Architecture in Fig.~\ref{fig: main_model})}]{
     \includegraphics[width=0.3\textwidth]{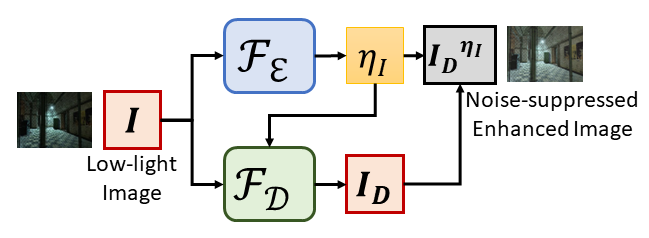}}
     \caption{The proposed low-light image enhancement approach, and its unpaired training and testing models.  While $\boldsymbol{I}$ and $\boldsymbol{W}$ respectively denote the unpaired input low-light and well-lit images, $\boldsymbol{I_D}$ and $\boldsymbol{W_D}$ respectively denote the outputs of noise-handling on $\boldsymbol{I}$ and $\boldsymbol{W}$. 
     $\boldsymbol{I}^{\alpha}$ denotes the partially enhanced image after the controlled transformation using $\alpha$, $\boldsymbol{I_D}^{\eta}$ denotes the enhanced image, and $\boldsymbol{I_D}^{\eta_I}$ represents the output noise-suppressed enhanced image. An $\eta_x$ with the subscript $x$ represents the enhancement map estimated from the input $x$ and $U$ stands for the all-ones matrix. $\mathcal{F_E}$ represents the enhancement module trained using the self-supervision loss $\mathcal{L_{SS}}$ and the self-conditioning losses $\mathcal{L_{SC}}$ and $\mathcal{L_{WSC}}$. $\mathcal{F_D}$ represents the noise-handling module trained using the self-conditioning loss $\mathcal{L_{DSC}}$, low-gradient magnitude suppression loss $\mathcal{L_G}$ and fidelity loss $\mathcal{L_{F}}$.}
     \label{fig:blockdia}
\vspace{-0.5 cm}
 \end{figure}

In this paper, we propose an unpaired deep learning approach for enhancing low-light images, which is based only on low-light image enhancement-specific attributes. Our deep network, which estimates the required degree of enhancement, is trained by leveraging novel self-supervision and self-conditioning strategies that work on unpaired low-light and well-lit images. The novel \textit{self-supervision strategy} considers only the input low-light image at hand and first applies a controlled transformation to enhance the image. Then, the self-supervised learning of our enhancement model is achieved by attempting to keep the overall enhancement of the input image by the model unaffected, in spite of the initial enhancement through the controlled transformation. The novel \textit{self-conditioning strategy} drives the learning of our enhancement model using the unpaired images. The learning via self-conditioning is performed by attempting not to enhance well-lit images that do not require enhancement and by avoiding further enhancement of low-light images that have been already enhanced using the model. The resilience to earlier controlled enhancements invoked by the self-supervision and the similarity in the self-conditioning using well-lit and already-enhanced images enable our model to learn appropriate required degrees of enhancement. While our self-supervision aims toward consistency in enhancement, our self-conditioning aims for its sufficiency.  Fig.~\ref{fig:blockdia} depicts our approach, whose components are discussed in Section~\ref{sec:algo}.

Further, our enhancement model contains a noise-handling module where the self-conditioning that noise-free well-lit images are not denoised is imposed for its training. Further, utilizing the fact that noise in low-light images is characterized by low gradient magnitude~\cite{li2018structure, dhara2021exposedness}, we regularize our noise-handling module using a relevant loss to suppress low gradient magnitude values. This is done in tandem with the maximization of fidelity between the input low-light and the output noise-suppressed low-light images targeting
detail preservation. The concerted use of self-conditioning and low-gradient magnitude suppression allows our model to learn latent representations that discriminate images with low-light specific noise from others enabling its handling. Several quantitative evaluations on multiple standard datasets having different varieties of low-light images show that our network mostly outperforms the relevant state-of-the-art while preserving details, quality, and naturalness. The subjective evaluation shows that our model enhances low-light images satisfactorily while handling the inherent noise. Ablation studies provided show the effectiveness of the network's low-light image enhancement-specific loss functions in executing the proposed self-supervision and self-conditioning strategies.

In a nutshell, the contributions of the paper are:
\begin{itemize}
    \item A low-light image enhancement network trained through unpaired deep learning using novel self-supervision and self-conditioning strategies.
    \item A novel self-supervision strategy based on a controlled transformation of the input image and a subsequent attempt to keep the image's enhancement consistent in spite of the transformation.
    \item A novel self-conditioning strategy that attempts to keep well-lit and already-enhanced images unenhanced when passed through the enhancement model.
    \item A novel detail-preserving noise-handling module based on self-conditioning and low gradient magnitude suppression to reduce low-light specific noise.
\end{itemize}

The rest of the paper is organized as follows. Section II
discusses the related work. Section III elaborately describes our proposed low-light image enhancement model, SelfEnNet. Section IV discusses the model framework and loss functions. Section V presents qualitative and quantitative comparisons of our approach with the state-of-the-art, and different studies on the proposed model. Finally, Section VI concludes the paper.

\section{Recent Related Work}
\label{sec:literature}
\subsection{Methods not using Learning Algorithms}
\label{subsec:literature_classical}
Most of the conventional low-light image enhancement techniques adopt the Retinex theory for decomposing the input image into reflectance and illuminance maps. 
Fu et al.~\cite{Fu2016} adopt a variational model of the Retinex framework for illumination map estimation. Guo et al.~\cite{Guo2017} adopt a structure prior in the Retinex framework and apply it on the maximum of R, G, and B channels to estimate the illumination map. 
Li et al.~\cite{li2018structure} introduce a noise term in the Retinex framework and estimate the noise to achieve denoising during the enhancement. Ren et al.~\cite{ren2020lr3m} inherently handle the noise through a noise-resilient estimation of the reflectance map using low-rank regularization. However, all these techniques perform a fixed gamma-based correction of the estimated illumination map for the required enhancement. Different from that, 
To improve the generalization performance of the deep model, a multi-scale architecture with illumination constraint has been proposed~\cite{fan2022multiscale}. %
Wang et al.~\cite{wang2019low} adopt an absorption light scattering model that estimates light scattering parameters to achieve the required enhancement. Dhara et al.~\cite{dhara2021exposedness}  perform enhancement based on exposedness quantification for visibility improvement. They consider an edge-preserving low-gradient magnitude suppression for handling noise. Liu et al.~\cite{liu2022efinet} propose an enhancement fusion mechanism in an iterative manner for low-light image enhancement. 
Recently, Singh et al.~\cite{singh2021principal} propose a principal component analysis (PCA) based fusion scheme that adopts a contrast enhancement technique for low-light image enhancement.

\subsection{Learning-based Methods}
\label{subsec:literature_deep}
The advent of deep learning and its success in image restoration lead to a plethora of deep learning based low light image enhancement techniques~\cite{9609683}. While Wei et al.~\cite{wei2018deep} demonstrate a Retinex based encoder-decoder framework for the enhancement, Zhao et al.~\cite{zhao2021retinexdip} present a Retinex based generative network for the same. 
Liu et al.~\cite{liu2021retinex} suggest a Retinex-inspired network that learns the strategy of finding prior architectures for enhancing images with different varieties of low-lighting conditions. The model also suppresses noise using a proposed noise removal module. Yang et al.~\cite{yang2021sparse} give a Retinex based model driven by sparse gradient constraints to learn a coupled representation in low-light and well-lit images. The learned representation is used for Retinex based decomposition to achieve the enhancement. Few works~\cite{guo2020zero, li2021learning, li2023pixel} present a deep model that estimates pixel-wise high-order curves for enhancement. 
Yang et al.~\cite{yang2020fidelity} suggest a semi-supervised two-stage network that exploits image fidelity and perceptual quality for the enhancement in a recursive way. A coarse-to-fine band representation in the model helps in handling the inherent noise. Zheng et al.~\cite{zheng2021adaptive} employ the total variation model in a deep framework for handling the inherent noise while enhancing the low-light images. Kar et al.~\cite{kar2021zero} exploit the Koschmieder light scattering model to propose a zero-shot approach for image restoration like low-light image enhancement. Tang et.al~\cite{9833451} adopt exposure information to develop a flexible low-light image enhancement technique. Lin et al.~\cite{9831058} develop a $L_p$-norm shrinkage mapping and an edge-preserving plug-and-play module to effectively estimate the illumination map for enhancement. Yucheng et al.~\cite{9656595} consider an enhanced V-channel conditioned data-driven network to achieve refined image enhancement.
Wenhui et al.~\cite{wu2022uretinex} present URetinex-Net that unfolds the Retinex framework for low-light image enhancement. Fu et al.~\cite{fu2023learning} propose a Retinex-based unsupervised enhancement technique using several novel reference-free losses. Guo et al.~\cite{guo2023low} present an illumination-guided enhancement technique that can handle complex degradation in low-light images. Luo et al.~\cite{luo2023pseudo} provide a framework that generates pseudo-well-lit images to eliminate the paired data dependency.
Ma et al.~\cite{ma2022toward} propose a self-calibrated framework for fast, flexible, and robust low-light image enhancement. Cui et al.~\cite{cui2022progressive} propose a progressive network with a dual branch that exploits the correlation and feature complementarity between the low-light image and its inverted version. Zhou et al.~\cite{zhou2022linear} propose a linear contrast enhancement network that consists of multiple subnets for adaptive brightness and linear contrast adjustment. Xu et al.~\cite{xu2022hfmnet} present a hierarchical feature mining network to preserve illumination and edge features of low-light images. Recently, Shi et al.~\cite{shi2023context} propose a lightweight model that addresses the overexposure issue using nonlocal context-based modeling.

\section{The Proposed Deep Model for Low-light Image Enhancement: \textit{SelfEnNet}}
\label{sec:algo}
The few recently proposed deep low-light image enhancement networks of \cite{guo2020zero, li2021learning, jiang2021enlightengan,lee2020unsupervised,zhang2020self,xiong2020unsupervised, zhao2021retinexdip, 9102962, liu2021retinex, kandula2023illumination}, which do not require ground truth data, have made such deep networks readily applicable to all kinds of real-world low-light images. They have not only ensured that the challenging requirement of registered low-light and well-lit images of the same real-world scene no longer exists but also have achieved performance comparable to that attained by networks trained on numerous input-ground truth image pairs.

However, immense scope for improvement remains that can be achieved by aiming for problem-specific optimality of the enhancement model by ensuring that only low-light image enhancement relevant attributes are leveraged and assumptions on image characteristics are avoided. While \cite{guo2020zero, li2021learning} assume a particular well-exposedness level and employs the gray-world assumption, \cite{jiang2021enlightengan, xiong2020unsupervised} use the generic adversarial loss on unpaired images. \cite{jiang2021enlightengan} also uses an illumination map for guidance, which inherently assumes a pre-defined image model. \cite{lee2020unsupervised} and \cite{kar2021zero} respectively use the predefined bright channel image prior and a low-light image formation model, and \cite{zhang2020self} assumes an entropic similarity between bright channels of the low-light and the enhanced images. \cite{zhao2021retinexdip, 9102962, liu2021retinex} consider generic smoothness constraints such as total variation as regularizers in their loss functions. At the same time, none of the models consider an explicit check on the sufficiency and consistency of the learned enhancement process.

Our proposed deep network model for low-light image enhancement is based on leveraging attributes specific to the enhancement for self-supervision and self-conditioning without any explicit or implicit assumption. While the self-supervision using the input image alone is targeted to achieve consistent enhancement, the self-conditioning using unpaired low-light and well-lit images is aimed to attain sufficient enhancement. A self-conditioning loss is also used to handle the low-light specific noise in the input image. Thus, our framework represents a completely distinct and novel approach of low-light image enhancement in comparison to existing unsupervised approaches like \cite{jiang2021enlightengan, guo2020zero}.

A detailed schematic representation of our model is shown in Fig.~\ref{fig:blockdia}. As can be seen from Fig.~\ref{fig:blockdia}(c), the model achieves enhancement through an estimated transformation from the input low-light image and the suppression of the inherent noise. Let us denote our low-light image enhancement model as $\mathcal{F}=\{\mathcal{F_E}, \mathcal{F_D}\}$, where $\mathcal{F_D}$ generates a denoised version of input low-light image and $\mathcal{F_E}$ generates a map of the required degree of enhancement (enhancement map) from the input low-light image that is used on the denoised low-light image to obtain the output enhanced image.

Let $\boldsymbol{I}$\footnote{Boldfaced variables represent multi-channel (color) entities} be the input low-light color image to our enhancement approach. We perform the enhancement as follows:
\begin{equation}
\boldsymbol{I_{E}}(p)=\boldsymbol{I_{D}}(p)^{\eta_{\boldsymbol{I}}(p)}
\label{eq: gamma}
\end{equation}
where $p$ is an image pixel location and 
$\boldsymbol{I_{E}}$ represents the enhanced image. $\eta_{\boldsymbol{I}}$ represents the enhancement map estimated for the image $\boldsymbol{I}$ using $\mathcal{F_E}$ on it. $\boldsymbol{I_{D}}$ represents the denoised low-light image obtained by subjecting $\boldsymbol{I}$ to $\mathcal{F_D}$, which uses $\eta_{\boldsymbol{I}}$. Then, $\eta_{\boldsymbol{I}}$ is applied pixel-wise on $\boldsymbol{I_{D}}$ to achieve the enhancement. The pixel-wise application of $\eta_{\boldsymbol{I}}$ on $\boldsymbol{I_{D}}$ allows the degree of enhancement to be different from one pixel to another, which enables our approach to avoid enhancing already well-lit regions of the input image and focus on enhancing the dark regions.

As mentioned in the introduction, the estimation of the map $\eta_{\boldsymbol{I}}$ is completely based on self-supervision and self-conditioning strategies leveraging low-light image enhancement relevant attributes. The estimation of $\boldsymbol{I_{D}}$ is based on low-light specific noise suppression involving another self-conditioning strategy. These strategies are used to train the entire model, in which the enhancement module $\mathcal{F_E}$ is trained first and then the noise-handling module $\mathcal{F_D}$ is trained.

\subsection{Enhancement Map Estimation}
\label{EMAPEstimation}
The enhancement map $\eta_{\boldsymbol{I}}$ used in (\ref{eq: gamma}) is obtained using $\mathcal{F_E}$ on $\boldsymbol{I}$, which is depicted as: 
\begin{equation}
    \eta_{\boldsymbol{I}}(p)=\mathcal{F_E}(\boldsymbol{I}), \forall p
    \label{eq: gammaest}
\end{equation}
We describe below the self-supervision via controlled transformation and the unpaired self-conditioning used to learn $\mathcal{F_E}$ for estimating $\eta_{\boldsymbol{I}}$ from $\boldsymbol{I}$.

\subsubsection{Self-supervision through controlled transformation}
\label{sec: Self-supervision_enh}

The self-supervision process driving the training of $\mathcal{F_E}$ uses only the input low-light image. From (\ref{eq: gammaest}), we have that by applying $\mathcal{F_E}$ on an input low-light image $\boldsymbol{I}$ we obtain $\eta_{\boldsymbol{I}}$. Let us rewrite it as:
\begin{equation}
    \eta_{\boldsymbol{I}}=\mathcal{F_E}(\boldsymbol{I})
    \label{eq:selfbeta}
\end{equation}
Further, we also obtain $\eta_{\boldsymbol{I_\alpha}}$ during the training as follows:
\begin{equation}
\eta_{\boldsymbol{I_\alpha}}=\mathcal{F_E}(\boldsymbol{I_\alpha})
    \label{eq:selfalpha1}
\end{equation}
where
\begin{equation}
\boldsymbol{I_\alpha}(p)=\boldsymbol{I}(p)^\alpha
    \label{eq:selfalpha2}
\end{equation}

(\ref{eq:selfalpha2}) depicts a controlled transformation, with $\alpha$ being an arbitrary positive constant less than $1$. Having obtained $\eta_{\boldsymbol{I}}$ and $\eta_{\boldsymbol{I_\alpha}}$ from (\ref{eq:selfbeta}) and (\ref{eq:selfalpha1}), respectively, our self-supervision strategy used in training $\mathcal{F_E}$ imposes the following constraint:
\begin{equation}
\eta_{\boldsymbol{I}}=\eta_{\boldsymbol{I_\alpha}} \times \alpha
\end{equation}
or equivalently:
\begin{equation}
    \mathcal{F_E}(\boldsymbol{I})=\mathcal{F_E}(\boldsymbol{I_\alpha}) \times \alpha
    \label{eq:selfsupcons}
\end{equation}
The above constraint targets consistency in the enhancement by attempting to keep the overall degree of enhancement for the image $\boldsymbol{I}$ at the same level ($\eta_{\boldsymbol{I}}$ estimated using $\mathcal{F_E}$ on $\boldsymbol{I}$) even when it is pre-enhanced through a controlled transformation (in (\ref{eq:selfalpha2})) of the same functional form as (\ref{eq: gamma}) using an arbitrary $\alpha$. As $\mathcal{F_E}$ is being constrained to estimate $\eta_{\boldsymbol{I_\alpha}}$ from $\boldsymbol{I_\alpha}$ if it estimates $\eta_{\boldsymbol{I}}$ from $\boldsymbol{I}$, a self-reference is being inherently invoked to estimate an appropriate $\eta_{\boldsymbol{I}}$. \color{black}Therefore, our self-supervision strategy can be perceived to be leveraging the following low-light image enhancement relevant attribute.
\color{black}
\textit{Attribute 1:} A low-light image enhancement model estimating an enhancement map $\eta_{\boldsymbol{I}}$ for an input image $\boldsymbol{I}$ should estimate the enhancement map $\eta_{\boldsymbol{I}}/\alpha$ when the input image is pre-enhanced using a controlled arbitrary enhancement map with all its elements as $\alpha$.

\subsubsection{Unpaired Self-conditioning}
\label{sec: self_condi_loss}
Our self-conditioning process involved in the training of $\mathcal{F_E}$ uses unpaired input low-light and well-lit images. Let us consider that during the model training, we obtain the map $\eta_{\boldsymbol{W}}$ by applying $\mathcal{F_E}$ on an input well-lit image $\boldsymbol{W}$, and as depicted earlier, the map $\eta_{\boldsymbol{I}}$ is produced by applying $\mathcal{F_E}$ on an input low-light image $\boldsymbol{I}$. Therefore, we have:
\begin{eqnarray}
\label{eq: betaagain}
\eta_{\boldsymbol{I}}&=&\mathcal{F_E}(\boldsymbol{I})\\
\eta_{\boldsymbol{W}}&=&\mathcal{F_E}(\boldsymbol{W})
\label{eq:well-lit}
\end{eqnarray}
where (\ref{eq: betaagain}) is rewritten from (\ref{eq: gammaest}). Further, we also obtain $\eta_{\boldsymbol{I_\omega}}$ during the training as follows:
\begin{equation}
\eta_{\boldsymbol{I_\omega}}=\mathcal{F_E}(\boldsymbol{I_\omega})
    \label{eq:well-en}
\end{equation}
where
\begin{equation}
    \boldsymbol{I_\omega}(p)=\boldsymbol{I}(p)^{\eta_{\boldsymbol{I}}(p)}
    \label{eq:11}
\end{equation}
In the above, $\boldsymbol{I_\omega}$ is the enhanced image obtained by applying the enhancement map $\eta_{\boldsymbol{I}}$ on the low-light image $\boldsymbol{I}$. Having obtained $\eta_{\boldsymbol{W}}$ and $\eta_{\boldsymbol{I_\omega}}$ from (\ref{eq:well-lit}) and (\ref{eq:well-en}), respectively, our self-conditioning strategy used in training $\mathcal{F_E}$ imposes the following two constraints:
\begin{eqnarray}
\eta_{\boldsymbol{W}} &=& U \\
\label{eq:selfcondconsA}
\eta_{\boldsymbol{I_\omega}} &=& U
\label{eq:selfcondconsB}
\end{eqnarray}
or equivalently:
\begin{equation}
    \mathcal{F_E}(\boldsymbol{W})=U\ \ \& \ \  \mathcal{F_E}(\boldsymbol{I_\omega})=U
    \label{eq:selfcondcons}
\end{equation}
where $U$ is a matrix with all elements as $1$. The above two constraints aim for sufficiency in the enhancement by attempting to keep the degrees of enhancement required for the already-enhanced image $\boldsymbol{I_\omega}$ and the well-lit image $\boldsymbol{W}$ as unity (no enhancement). As $\mathcal{F_E}$ is being constrained to avoid enhancing an already-enhanced or a well-lit image in a similar manner, a self-reference is being essentially used that checks the sufficiency of the enhancement already performed by $\mathcal{F_E}$ on the low-light image $\boldsymbol{I}$, where the sufficiency is guided by the well-lit images. \color{black}Therefore, our self-conditioning strategy can be interpreted to be exploiting the following two low-light image enhancement specific attributes.\color{black}

\textit{Attribute 2:}
A low-light image enhancement model should not enhance an image $\boldsymbol{I_\omega}$ obtained by enhancing an input low-light image $\boldsymbol{I}$ using the same model.

\textit{Attribute 3:}
A low-light image enhancement model should not enhance a well-lit input image $\boldsymbol{W}$.

While the constraint in (\ref{eq:selfsupcons}) of our self-supervision strategy is implemented using the loss function $\mathcal{L_{SS}}$ discussed in Section~\ref{SSloss}, the constraints in (\ref{eq:selfcondcons}) of our self-conditioning strategy is implemented using the losses $\mathcal{L_{SC}}$ and $\mathcal{L_{WSC}}$ described in Sections~\ref{LSCloss} and~\ref{WSCloss}, respectively. Therefore, the overall loss function employed to train $\mathcal{F_E}$ is:
\begin{eqnarray}
    1[\textrm{arg}(\mathcal{F_E})=\boldsymbol{I}]\big{(}\mathcal{L_{SS}}(\boldsymbol{I},\alpha)+c_1\mathcal{L_{SC}}(\boldsymbol{I_\omega})\big{)}\nonumber\\+1[\textrm{arg}(\mathcal{F_E})=\boldsymbol{W}]\big{(}c_2\mathcal{L_{WSC}}(\boldsymbol{W})\big{)}
\end{eqnarray}
where $1[\mathcal{C}]$ takes the value $1$ when the condition $\mathcal{C}$ is true, otherwise it takes the value $0$. We take $c_1=c_2=10^{-2}$. The related ablation study is provided in Section~\ref{sec:ELoss}.

\subsection{Low-light Specific Noise-handling}
\label{Noise}
Low-light images are prone to noise, and we have a separate module $\mathcal{F_D}$ in our enhancement model to handle it. As depicted in (\ref{eq: gamma}), the enhancement map $\eta_{\boldsymbol{I}}$ for an input low-light image $\boldsymbol{I}$ obtained using $\mathcal{F_E}$ is applied on the noise suppressed version $\boldsymbol{I_D}$ of the input image. We use $\mathcal{F_D}$ to produce $\boldsymbol{I_D}$ from $\boldsymbol{I}$, which is depicted as:
\begin{equation}
    \boldsymbol{I_D}(p)=\mathcal{F_D}(\boldsymbol{I}, \eta_{\boldsymbol{I}}), \forall p
\end{equation}
where $\mathcal{F_D}$ utilizes the knowledge of $\eta_{\boldsymbol{I}}$ obtained using the already-trained $\mathcal{F_E}$ on $\boldsymbol{I}$. The map $\eta_{\boldsymbol{I}}$ indicates the different degrees of enhancement required in different image regions, allowing emphasis by $\mathcal{F_D}$ on the low-lit areas of the image $\boldsymbol{I}$.

Noise in low-light images is image-dependent \cite{wei2020physics} and is usually characterized by low gradient magnitude values \cite{li2018structure, dhara2021exposedness}. Hence, it is imperative that an appropriate $\mathcal{F_D}$ should consider low gradient magnitude suppression for the denoising. At the same time,  $\mathcal{F_D}$ should aim at preserving the remaining image details, which will then be acted upon by $\eta_{\boldsymbol{I}}$ as in (\ref{eq: gamma}) for the enhancement. Therefore, we consider the minimization of the low gradient magnitude content $\mathcal{L_{G}}$ in $I_D$ along with the fidelity loss $\mathcal{L_{F}}$ between $I_D$ and $I$. So, the minimization objective is:
\begin{equation}
    \mathcal{L_{F}}(\boldsymbol{I_D},\boldsymbol{I}) + c_3\mathcal{L_{G}}(\boldsymbol{I_D})
    \label{eq:fidLgrad}
\end{equation}
where $\mathcal{L_{G}}$ and $\mathcal{L_{F}}$ used in the training of $\mathcal{F_D}$ are discussed in Section~\ref{GLoss} and  Section~\ref{FLoss}, respectively. Note that as $I_D$ is computed from the input low-light image $I$ using $\mathcal{F_D}$, only the input image is required in above learning process.

Further, the image-dependent nature of the noise in low-light images suggests the requirement of a strategy that makes the latent representations in $\mathcal{F_D}$ to be discriminative between images with low-light specific noise and images without such noise. We achieve this by imposing a self-conditioning strategy on $\mathcal{F_D}$.

\subsubsection{Self-conditioning for denoising}
\label{sec: denoise_self_condi}

We devise our self-conditioning strategy based on the fact that while $\mathcal{F_D}$ minimizes low gradient magnitude and fidelity loss to achieve noise reduction when working on a low-light image $\boldsymbol{I}$, it does not require to perform the same when working on a well-lit image $\boldsymbol{W}$. This is because $\boldsymbol{W}$ is not expected to contain low-light specific noise and a well-lit image may not be enhanced at all in (\ref{eq: gamma}) due to (\ref{eq:selfcondconsA}). Therefore, we employ the following self-conditioning loss using a well-lit image $\boldsymbol{W}$ that is minimized during the training of $\mathcal{F_D}$:
\begin{equation}
    \mathcal{L_{DSC}}(\boldsymbol{W_D},\boldsymbol{W})
    \label{eq:fidW}
\end{equation}
where
\begin{equation}
\boldsymbol{W_D}(p)=\mathcal{F_D}(\boldsymbol{W}, \eta_{\boldsymbol{W}}), \forall p
\end{equation}
with $\eta_{\boldsymbol{W}}$ obtained using the already-trained $\mathcal{F_E}$ on $\boldsymbol{W}$. The above self-conditioning loss $\mathcal{L_{DSC}}$ is described in Section~\ref{DSelfLoss}. The use of $\mathcal{L_{DSC}}$ to train $\mathcal{F_D}$ empowers it with the capability to differentiate embedded low-light specific noise from other image contents and details. Therefore, the use of $\mathcal{L_{DSC}}$ along with $\mathcal{L_{G}}$ and $\mathcal{L_{F}}$ from (\ref{eq:fidLgrad})
allows $\mathcal{F_D}$ to specifically be effective with low-light relevant noise rather than with any generic image detail. So, the overall loss function employed to train $\mathcal{F_D}$ is:
\begin{eqnarray}
    1[\textrm{arg}(\mathcal{F_D})=\boldsymbol{I}]\bigg{(}c_4\Big{(} \mathcal{L_{F}}(\boldsymbol{I_D},\boldsymbol{I}) + c_3\mathcal{L_{G}}(\boldsymbol{I_D}) \Big{)}\bigg{)}\nonumber\\+1[\textrm{arg}(\mathcal{F_D})=\boldsymbol{W}]\Big{(}\mathcal{L_{DSC}}(\boldsymbol{W_D},\boldsymbol{W})\Big{)}
\end{eqnarray}
where we take $c_3=c_4=10^{-1}$. The related ablation study is provided in Section~\ref{sec:DLoss}.

\begin{figure*}
    \centering
    \includegraphics[width=\textwidth]{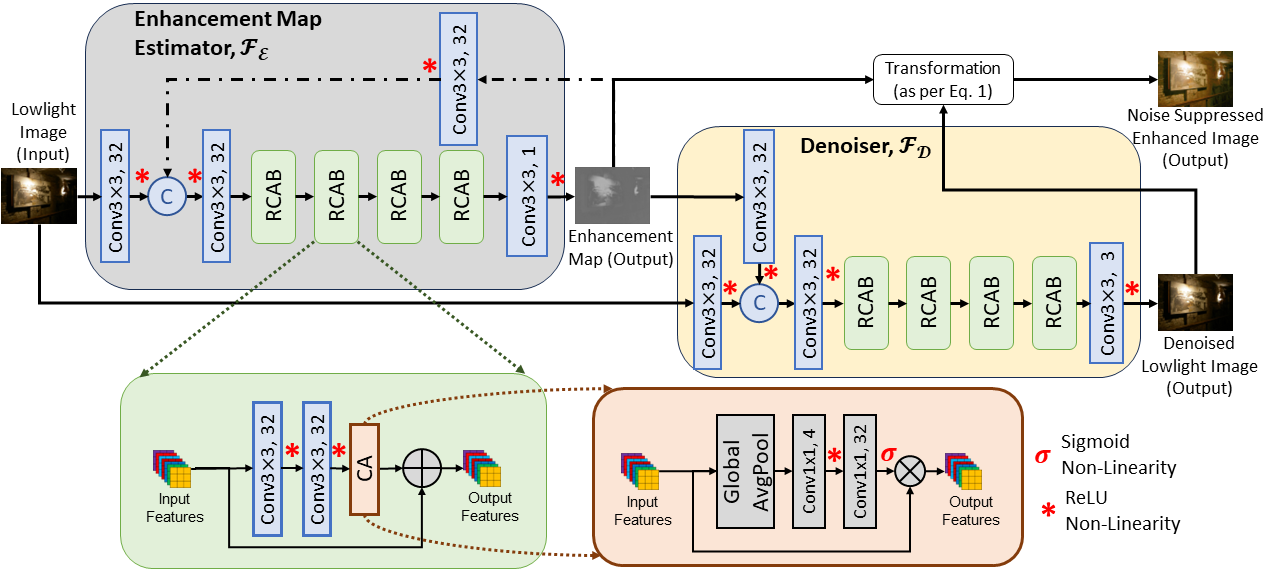}
    \caption{\color{black} The detailed architecture of our low-light image enhancement network with the enhancement $\mathcal{F_E}$ and noise-handling $\mathcal{F_D}$ modules.}
    \label{fig: main_model}
\end{figure*}

\subsection{Loss Functions}
The loss functions used to impose our self-supervision and self-conditioning of the enhancement network $\mathcal{F_E}$, and the self-conditioning of the denoising network $\mathcal{F_D}$ is discussed here. We also describe here the fidelity loss and the low gradient magnitude related loss used on $\mathcal{F_D}$.

\subsubsection{Enhancement Self-supervision Loss} 
\label{SSloss}

\color{black}
To invoke the proposed novel self-supervision strategy through controlled transformation as explained in Section~\ref{sec: Self-supervision_enh}, we introduce the enhancement self-supervision loss. \color{black} This loss which uses a low-light image $\boldsymbol{I}$ and its arbitrarily enhanced version $\boldsymbol{I_\alpha}$ (by $\mathcal{F_E}$ using $\alpha$) in (\ref{eq:selfsupcons}) to train $\mathcal{F_E}$ is implemented based on the $L_2$ loss expression as follows:
\begin{equation}
\mathcal{L_{SS}}(\boldsymbol{I},\alpha)=
||\mathcal{F_E}(\boldsymbol{I})-(\mathcal{F_E}(\boldsymbol{I_\alpha}) \times \alpha)||_2^2
\end{equation}
where we empirically choose $\alpha=0.75$. An ablation study related to this is provided in Table \ref{tab:self_enc}.

\subsubsection{Enhancement Self-conditioning Loss} 
\label{LSCloss}

\color{black}
The proposed unpaired self-conditioning strategy explained in Section~\ref{sec: self_condi_loss} is implemented by introducing the enhancement self-conditioning loss.
\color{black}
The loss which uses an enhanced (by $\mathcal{F_E}$) low-light image $\boldsymbol{I_\omega}$ in (\ref{eq:selfcondcons}) to train $\mathcal{F_E}$ is implemented using the $L_2$ loss expression as:
\begin{equation}
\mathcal{L_{SC}}(\boldsymbol{I_\omega})=||\mathcal{F_E}(\boldsymbol{I_\omega})-U||_2^2
\end{equation}
where, as said earlier, $U$ is a matrix with all elements as $1$.

\subsubsection{Well-lit Self-conditioning Loss}
\label{WSCloss}

\color{black}
Conditioning from well-lit images is also a part of the unpaired self-conditioning strategy as explained in Section~\ref{sec: self_condi_loss}, which is implemented using a well-lit self-conditioning loss.
\color{black}
This self-conditioning loss that uses a well-lit image $\boldsymbol{W}$ in (\ref{eq:selfcondcons}) to train $\mathcal{F_E}$ is implemented using the $L_2$ loss expression again as:
\begin{equation}
\mathcal{L_{WSC}}(\boldsymbol{W})=||\mathcal{F_E}(\boldsymbol{W})-U||_2^2
\end{equation}
\subsubsection{Low Gradient Magnitude Related Loss}
\label{GLoss}

\color{black}
Low gradient magnitude suppression for denoising, as discussed in Section~\ref{Noise}, is invoked through 
\color{black}
a low gradient magnitude related loss. This loss that is used on a denoised low-light image $\boldsymbol{I_D}$ (by $\mathcal{F_D}$) in (\ref{eq:fidLgrad}) to train $\mathcal{F_D}$ is implemented as:
\begin{equation}
\mathcal{L_{G}}(\boldsymbol{I_D})={w_{x,p}}(\boldsymbol{I_D}){\left( {\frac{{\partial \boldsymbol{I_D}}}{{\partial x}}} \right)_p^2} + {w_{y,p}}(\boldsymbol{I_D}){\left( {\frac{{\partial \boldsymbol{I_D}}}{{\partial y}}} \right)_p^2}
\end{equation}
where ${\left( {\frac{{\partial I}}{{\partial x}}} \right)_p}$ and ${\left( {\frac{{\partial I}}{{\partial y}}} \right)_p}$ represent the x-direction and y-direction gradients respectively at the pixel location $p$. The corresponding weights $w_{x,p}$ and $w_{y,p}$ are inversely related to the corresponding gradients to ensure emphasis on low gradient values in the computation. These weights are determined as defined in ~\cite{farbman2008edge,lischinski2006interactive}: 
\begin{equation}
{w_{x,p}}(\boldsymbol{I_D}) = {\left( {{{\left| {\frac{{\partial \ell }}{{\partial x}}} \right|}^\delta_p} + \varepsilon } \right)^{ - 1}}{w_{y,p}}(g) = {\left( {{{\left| {\frac{{\partial \ell }}{{\partial y}}} \right|}^\delta_p} + \varepsilon } \right)^{ - 1}}
\end{equation} 
where $l$ is the log-luminance channel of $\boldsymbol{I_D}$ and $\delta=1.2$ \cite{farbman2008edge} is considered.

\subsubsection{Fidelity Loss}
\label{FLoss}

\color{black}
To preserve the structural information in the low-light image during denoising, a fidelity loss is used in the training.
\color{black}
The pixel-level fidelity loss between a low-light image $\boldsymbol{I}$ and its denoised version $\boldsymbol{I_D}$ (by $\mathcal{F_D}$) used in (\ref{eq:fidLgrad}) to train $\mathcal{F_D}$ is implemented using the $L_2$ loss expression as:
\begin{equation}
    \mathcal{L_{F}}(\boldsymbol{I_D},\boldsymbol{I})=||\boldsymbol{I_D}-\boldsymbol{I}||_2^2=||\mathcal{F_D}(\boldsymbol{I})-\boldsymbol{I}||_2^2
\end{equation}
{$\mathcal{F_N}$  takes an input as low-light image $I$ to produce a denoised version $\boldsymbol{I_D}$ that is $\boldsymbol{I_D}=\mathcal{F_N}(\boldsymbol{I_G})$. To ensure detail-aware denoising, we define the following L2 loss function, which is nothing but a strategy for fidelity preservation.}

\subsubsection{Denoising Self-conditioning Loss}
\label{DSelfLoss}
\color{black}
The novel self-conditioning during denoising discussed in Section~\ref{sec: denoise_self_condi} is invoked using the denoising self-conditioning loss.
\color{black}
This loss between a well-lit image $\boldsymbol{W}$ and its denoised version $\boldsymbol{W_D}$ (by $\mathcal{F_D}$) used in (\ref{eq:fidW}) to train $\mathcal{F_D}$ is implemented using the $L_2$ loss expression again as:
\begin{equation}
    \mathcal{L_{DSC}}(\boldsymbol{W_D},\boldsymbol{W})=||\boldsymbol{W_D}-\boldsymbol{W}||_2^2=||\mathcal{F_D}(\boldsymbol{W})-\boldsymbol{W}||_2^2
\end{equation}

\definecolor{DGreen}{rgb}{0.0, 0.5, 0.0}
\definecolor{DRed}{rgb}{0.9, 0.17, 0.31}

\begin{table*}[]
\caption{Performance comparison on the $100$ real-world test images from the latest LOL dataset~\cite{yang2021sparse}. {\textbf{Bold}}: Best without paired supervison, \underline{Underline}: Best Overall}
\label{tab: LO_Table}
\label{tab: LOL}
\resizebox{\textwidth}{!}{
\begin{tabular}{|l|lllll|llllllllll|}
\hline
\multirow{3}{*}{Measures} &
  \multicolumn{5}{c|}{\multirow{2}{*}{Paired Supervision}} &
  \multicolumn{10}{c|}{No paired Supervision} \\ \cline{7-16} 
 &
  \multicolumn{5}{c|}{} &
  \multicolumn{8}{c|}{No Unpaired Supervision} &
  \multicolumn{2}{l|}{Unpaired Supervision} \\ \cline{2-16} 
 &
 \multicolumn{1}{l|}{\begin{tabular}[c]{@{}l@{}}DRD~\cite{wei2018deep}\\ BMVC'21\end{tabular}} & \multicolumn{1}{l|}{\begin{tabular}[c]{@{}l@{}}DRBN~\cite{yang2021band}\\ TIP'21\end{tabular}}&
\multicolumn{1}{l|}{\begin{tabular}[c]{@{}l@{}}URN~\cite{wu2022uretinex}\\ CVPR'22\end{tabular}}&
\multicolumn{1}{l|}{\begin{tabular}[c]{@{}l@{}}MLLEN~\cite{fan2022multiscale}\\ TCSVT'22\end{tabular}}&
  \begin{tabular}[c]{@{}l@{}}Bread~\cite{guo2023low}\\ IJCV'23\end{tabular}  &
  \multicolumn{1}{l|}{\begin{tabular}[c]{@{}l@{}}LIME~\cite{Guo2017}\\ TIP'17\end{tabular}} &
  \multicolumn{1}{l|}{\begin{tabular}[c]{@{}l@{}}RRM~\cite{li2018structure}\\ TIP'18\end{tabular}} &
  \multicolumn{1}{l|}{\begin{tabular}[c]{@{}l@{}}ALSM~\cite{wang2019low}\\ TIP'19\end{tabular}} &
  \multicolumn{1}{l|}{\begin{tabular}[c]{@{}l@{}}Zero-DCE~\cite{guo2020zero}\\ CVPR'20\end{tabular}} &
  \multicolumn{1}{l|}{\begin{tabular}[c]{@{}l@{}}ZCP$^*$~\cite{kar2021zero}\\ CVPR'21\end{tabular}} &
  \multicolumn{1}{l|}{\begin{tabular}[c]{@{}l@{}}RUAS~\cite{liu2021retinex}\\ CVPR'21\end{tabular}} &
  \multicolumn{1}{l|}{\begin{tabular}[c]{@{}l@{}}SCI~\cite{ma2022toward}\\ CVPR'22\end{tabular}} &
  \multicolumn{1}{l|}{\begin{tabular}[c]{@{}l@{}}PairLIE~\cite{fu2023learning}\\ CVPR'23\end{tabular}} &
  \multicolumn{1}{l|}{\begin{tabular}[c]{@{}l@{}}EGAN~\cite{jiang2021enlightengan}\\ TIP'21\end{tabular}} &
  \multicolumn{1}{l|}{\begin{tabular}[c]{@{}l@{}}SelfEnNet\\ (Ours)\end{tabular}} \\ \hline
  \begin{tabular}[c]{@{}l@{}}PSNR\\ SSIM\\ CIEDE\\LPIPS$_\textrm{VGG}$\end{tabular} &
  \multicolumn{1}{l|}{\begin{tabular}[c]{@{}l@{}}16.09\\ 0.46\\ 49.40\\0.47\\\end{tabular}} &
  \multicolumn{1}{l|}{\begin{tabular}[c]{@{}l@{}}{20.19}\\  {0.82}\\  {38.05}\\{0.27}\\\end{tabular}} &
  \multicolumn{1}{l|}{\begin{tabular}[c]{@{}l@{}}{21.09}\\  {0.83}\\  {37.11}\\{\underline{0.21}}\\\end{tabular}} &
  \multicolumn{1}{l|}{\begin{tabular}[c]{@{}l@{}}{17.46}\\  {0.49}\\  {55.53}\\{0.43}\\\end{tabular}} &
  \begin{tabular}[c]{@{}l@{}} {\underline{23.69}}\\  {\underline{0.86}}\\  {\underline{31.23}}\\{0.27}\end{tabular} &
  \multicolumn{1}{l|}{\begin{tabular}[c]{@{}l@{}}17.78\\ 0.54\\ 50.08\\0.38\end{tabular}} &
  \multicolumn{1}{l|}{\begin{tabular}[c]{@{}l@{}}17.34\\ 0.69\\ 45.42\\0.36\end{tabular}} &
  \multicolumn{1}{l|}{\begin{tabular}[c]{@{}l@{}}16.71\\ 0.48\\ 60.79\\0.39\end{tabular}} &
  \multicolumn{1}{l|}{\begin{tabular}[c]{@{}l@{}}18.05\\ 0.60\\ 52.77\\0.36\end{tabular}} &
   \multicolumn{1}{l|}{\begin{tabular}[c]{@{}l@{}}16.42\\ 0.57\\ 46.90\\0.41\end{tabular}} &
  \multicolumn{1}{l|}{\begin{tabular}[c]{@{}l@{}}15.32\\ 0.51\\ 60.34\\0.41\end{tabular}} &
    \multicolumn{1}{l|}{\begin{tabular}[c]{@{}l@{}}17.30\\ 0.55\\ 58.47\\0.35\end{tabular}} &
    \multicolumn{1}{l|}{\begin{tabular}[c]{@{}l@{}}19.88\\ 0.74\\ 42.79\\0.34\end{tabular}} &
  \multicolumn{1}{l|}{\begin{tabular}[c]{@{}l@{}}18.63\\ 0.63\\ 49.62\\0.38\end{tabular}} &
 {\begin{tabular}[c]{@{}l@{}} {{\textbf{21.36}}}\\  {\textbf{0.77}}\\  {\textbf{39.43}}\\ {\textbf{0.31}}\end{tabular}} \\ \hline 
\end{tabular}}
$^*$An image restoration approach targeted for dehazing, whose use in low-light image enhancement was also demonstrated
\vspace{-0.3 cm}
\end{table*}

\begin{table*}[]
\caption{Comparison on $5$ standard low-light image datasets using well-accepted no-reference measures. {\color{DGreen}Green}: Best, {\color{DRed}Red}: Second Best.}
\label{tab: Real}
\resizebox{\textwidth}{!}{
\begin{tabular}{|l|l|lllll|lllllllll|}
\hline
\multirow{3}{*}{Dataset} &
  \multirow{3}{*}{Measures} &
  \multicolumn{5}{c|}{\multirow{2}{*}{Paired Supervision}} &
  \multicolumn{9}{c|}{No paired Supervision} \\ \cline{8-16} 
 &
   &
  \multicolumn{5}{c|}{} &
  \multicolumn{7}{c|}{No Unpaired Supervision} &
  \multicolumn{2}{l|}{Unpaired Supervision} \\ \cline{3-16} 
 &
   &
  \multicolumn{1}{l|}{\begin{tabular}[c]{@{}l@{}}DRD~\cite{wei2018deep}\\ BMVC'21\end{tabular}} &  \multicolumn{1}{l|}{\begin{tabular}[c]{@{}l@{}}DRBN~\cite{yang2021band}\\ TIP'21\end{tabular}} &
  \multicolumn{1}{l|}{\begin{tabular}[c]{@{}l@{}}URN~\cite{wu2022uretinex}\\ CVPR'22\end{tabular}} &
  \multicolumn{1}{l|}{\begin{tabular}[c]{@{}l@{}}MLLEN~\cite{fan2022multiscale}\\ TCSVT'22\end{tabular}} &
  \begin{tabular}[c]{@{}l@{}}Bread~\cite{guo2023low}\\ IJCV'23\end{tabular}  &
  \multicolumn{1}{l|}{\begin{tabular}[c]{@{}l@{}}LIME~\cite{Guo2017}\\ TIP'17\end{tabular}} &
  \multicolumn{1}{l|}{\begin{tabular}[c]{@{}l@{}}RRM~\cite{li2018structure}\\ TIP'18\end{tabular}} &
  \multicolumn{1}{l|}{\begin{tabular}[c]{@{}l@{}}ALSM~\cite{wang2019low}\\ TIP'19\end{tabular}} &
  \multicolumn{1}{l|}{\begin{tabular}[c]{@{}l@{}}Zero-DCE~\cite{guo2020zero}\\ CVPR'20\end{tabular}} &
  \multicolumn{1}{l|}{\begin{tabular}[c]{@{}l@{}}RUAS~\cite{liu2021retinex}\\ CVPR'21\end{tabular}} &
  \multicolumn{1}{l|}{\begin{tabular}[c]{@{}l@{}}SCI~\cite{ma2022toward}\\ CVPR'22\end{tabular}} &
  \multicolumn{1}{l|}{\begin{tabular}[c]{@{}l@{}}PairLIE~\cite{fu2023learning}\\ CVPR'23\end{tabular}} &
  \multicolumn{1}{l|}{\begin{tabular}[c]{@{}l@{}}EGAN~\cite{jiang2021enlightengan}\\ TIP'21\end{tabular}} &
  \multicolumn{1}{l|}{\begin{tabular}[c]{@{}l@{}}SelfEnNet\\ (Ours)\end{tabular}} \\ \hline
 VV &
  \begin{tabular}[c]{@{}l@{}}NIQE\\ LOE\end{tabular} &
  \multicolumn{1}{l|}{\begin{tabular}[c]{@{}l@{}}2.61\\  391.74\end{tabular}} & \multicolumn{1}{l|}{\begin{tabular}[c]{@{}l@{}}2.59\\ 409.76\end{tabular}}&
  \multicolumn{1}{l|}{\begin{tabular}[c]{@{}l@{}}2.37\\ 169.43\end{tabular}}&
  \multicolumn{1}{l|}{\begin{tabular}[c]{@{}l@{}}{\color{DRed}2.17}\\ 595.56\end{tabular}}&
  \begin{tabular}[c]{@{}l@{}}2.48\\ 203.79\end{tabular} &
  \multicolumn{1}{l|}{\begin{tabular}[c]{@{}l@{}}2.44\\ 460.67\end{tabular}} &
  \multicolumn{1}{l|}{\begin{tabular}[c]{@{}l@{}}2.64\\ 254.26\end{tabular}} &
  \multicolumn{1}{l|}{\begin{tabular}[c]{@{}l@{}}2.55\\ 391.56\end{tabular}} &
  \multicolumn{1}{l|}{\begin{tabular}[c]{@{}l@{}} {\color{DRed}2.17}\\  {\color{DRed}99.49}\end{tabular}} &
  \multicolumn{1}{l|}{\begin{tabular}[c]{@{}l@{}}3.81\\ 583.70\end{tabular}} &
  \multicolumn{1}{l|}{\begin{tabular}[c]{@{}l@{}} {2.30}\\  {108.97}\end{tabular}} &
  \multicolumn{1}{l|}{\begin{tabular}[c]{@{}l@{}} {2.91}\\  {142.41}\end{tabular}} &
  \multicolumn{1}{l|}{\begin{tabular}[c]{@{}l@{}}3.45\\ 464.38\end{tabular}} &
  \begin{tabular}[c]{@{}l@{}} {\color{DGreen}2.07}\\  {\color{DGreen}51.78}\end{tabular} \\ \hline
NPE &
  \begin{tabular}[c]{@{}l@{}}NIQE\\ LOE\end{tabular} &
  \multicolumn{1}{l|}{\begin{tabular}[c]{@{}l@{}}4.06\\ 653.84\end{tabular}} & \multicolumn{1}{l|}{\begin{tabular}[c]{@{}l@{}}3.57\\ 391.54\end{tabular}}&
  \multicolumn{1}{l|}{\begin{tabular}[c]{@{}l@{}}4.05\\ 405.17\end{tabular}}&
  \multicolumn{1}{l|}{\begin{tabular}[c]{@{}l@{}}3.41\\ 544.97\end{tabular}}&
  \begin{tabular}[c]{@{}l@{}}3.44\\ 378.75\end{tabular} &
  \multicolumn{1}{l|}{\begin{tabular}[c]{@{}l@{}}3.79\\ 823.00\end{tabular}} &
  \multicolumn{1}{l|}{\begin{tabular}[c]{@{}l@{}}3.97\\ 633.56\end{tabular}} &
  \multicolumn{1}{l|}{\begin{tabular}[c]{@{}l@{}}3.75\\ 972.48\end{tabular}} &
  \multicolumn{1}{l|}{\begin{tabular}[c]{@{}l@{}}3.47\\  {\color{DRed}161.40}\end{tabular}} &
  \multicolumn{1}{l|}{\begin{tabular}[c]{@{}l@{}}6.38\\ 1357.49\end{tabular}} &
  \multicolumn{1}{l|}{\begin{tabular}[c]{@{}l@{}}4.16\\  {359.79}\end{tabular}} &
  \multicolumn{1}{l|}{\begin{tabular}[c]{@{}l@{}}3.48\\  {377.88}\end{tabular}} &
  \multicolumn{1}{l|}{\begin{tabular}[c]{@{}l@{}} {\color{DRed}3.33}\\ 749.14\end{tabular}} &
  \begin{tabular}[c]{@{}l@{}} {\color{DGreen}3.22}\\  {\color{DGreen}72.84}\end{tabular} \\ \hline
LIME &
  \begin{tabular}[c]{@{}l@{}}NIQE\\ LOE\end{tabular} &
  \multicolumn{1}{l|}{\begin{tabular}[c]{@{}l@{}}4.90\\ 539.64\end{tabular}} & \multicolumn{1}{l|}{\begin{tabular}[c]{@{}l@{}}3.86\\ 458.82\end{tabular}}&
  \multicolumn{1}{l|}{\begin{tabular}[c]{@{}l@{}}4.09\\ 201.72\end{tabular}}&
  \multicolumn{1}{l|}{\begin{tabular}[c]{@{}l@{}}3.57\\ 344.42\end{tabular}}&
  \begin{tabular}[c]{@{}l@{}}4.13\\ 463.58\end{tabular} &
  \multicolumn{1}{l|}{\begin{tabular}[c]{@{}l@{}}4.19\\ 559.61\end{tabular}} &
  \multicolumn{1}{l|}{\begin{tabular}[c]{@{}l@{}}3.90\\ 276.12\end{tabular}} &
  \multicolumn{1}{l|}{\begin{tabular}[c]{@{}l@{}}4.15\\ 495.05\end{tabular}} &
  \multicolumn{1}{l|}{\begin{tabular}[c]{@{}l@{}}3.78\\  {135.03}\end{tabular}} &
  \multicolumn{1}{l|}{\begin{tabular}[c]{@{}l@{}}4.26\\ 719.90\end{tabular}} &
  \multicolumn{1}{l|}{\begin{tabular}[c]{@{}l@{}}4.14\\  {\color{DRed}75.49}\end{tabular}} &
  \multicolumn{1}{l|}{\begin{tabular}[c]{@{}l@{}}4.31\\  {241.15}\end{tabular}} &
  \multicolumn{1}{l|}{\begin{tabular}[c]{@{}l@{}} {\color{DGreen}3.37}\\ 540.62\end{tabular}} &
  \begin{tabular}[c]{@{}l@{}} {\color{DRed}3.40}\\  {\color{DGreen}68.37}\end{tabular} \\ \hline
Fusion &
  \begin{tabular}[c]{@{}l@{}}NIQE\\ LOE\end{tabular} &
  \multicolumn{1}{l|}{\begin{tabular}[c]{@{}l@{}}3.99\\ 504.31\end{tabular}} &\multicolumn{1}{l|}{\begin{tabular}[c]{@{}l@{}}3.17\\ 319.97\end{tabular}}&
  \multicolumn{1}{l|}{\begin{tabular}[c]{@{}l@{}}3.46\\ 295.18\end{tabular}}&
  \multicolumn{1}{l|}{\begin{tabular}[c]{@{}l@{}}2.87\\ 395.71\end{tabular}}&
  \begin{tabular}[c]{@{}l@{}}3.23\\ 287.18\end{tabular} &
  \multicolumn{1}{l|}{\begin{tabular}[c]{@{}l@{}}3.50\\ 680.74\end{tabular}} &
  \multicolumn{1}{l|}{\begin{tabular}[c]{@{}l@{}}3.94\\ 408.39\end{tabular}} &
  \multicolumn{1}{l|}{\begin{tabular}[c]{@{}l@{}}3.54\\ 688.82\end{tabular}} &
  \multicolumn{1}{l|}{\begin{tabular}[c]{@{}l@{}}3.02\\  {\color{DRed}139.54}\end{tabular}} &
  \multicolumn{1}{l|}{\begin{tabular}[c]{@{}l@{}}5.07\\ 1014.37\end{tabular}} &
  \multicolumn{1}{l|}{\begin{tabular}[c]{@{}l@{}}3.79\\  {221.41}\end{tabular}} &
  \multicolumn{1}{l|}{\begin{tabular}[c]{@{}l@{}}3.75\\  {278.21}\end{tabular}} &
  \multicolumn{1}{l|}{\begin{tabular}[c]{@{}l@{}} {\color{DGreen}2.78}\\ 499.54\end{tabular}} &
  \begin{tabular}[c]{@{}l@{}} {\color{DRed}2.80}\\  {\color{DGreen}56.28}\end{tabular} \\ \hline
DICM &
  \begin{tabular}[c]{@{}l@{}}NIQE\\ LOE\end{tabular} &
  \multicolumn{1}{l|}{\begin{tabular}[c]{@{}l@{}}4.30\\ 636.16\end{tabular}} & \multicolumn{1}{l|}{\begin{tabular}[c]{@{}l@{}}3.37\\ 454.60\end{tabular}}&
  \multicolumn{1}{l|}{\begin{tabular}[c]{@{}l@{}}3.40\\ 480.41\end{tabular}}&
  \multicolumn{1}{l|}{\begin{tabular}[c]{@{}l@{}}{\color{DGreen}2.96}\\ 399.05\end{tabular}}&
  \begin{tabular}[c]{@{}l@{}}3.41\\ 341.21\end{tabular} & 
  \multicolumn{1}{l|}{\begin{tabular}[c]{@{}l@{}} {\color{DRed}3.05}\\ 818.61\end{tabular}} &
  \multicolumn{1}{l|}{\begin{tabular}[c]{@{}l@{}}3.37\\ 541.97\end{tabular}} &
  \multicolumn{1}{l|}{\begin{tabular}[c]{@{}l@{}} {\color{DGreen}2.96}\\ 865.63\end{tabular}} &
  \multicolumn{1}{l|}{\begin{tabular}[c]{@{}l@{}}3.43\\  {\color{DRed}232.50}\end{tabular}} &
  \multicolumn{1}{l|}{\begin{tabular}[c]{@{}l@{}}4.89\\ 1421.42\end{tabular}} &
  \multicolumn{1}{l|}{\begin{tabular}[c]{@{}l@{}}3.61\\  {321.87}\end{tabular}} &
  \multicolumn{1}{l|}{\begin{tabular}[c]{@{}l@{}}3.79\\  {375.37}\end{tabular}} &
  \multicolumn{1}{l|}{\begin{tabular}[c]{@{}l@{}} {\color{DRed}3.05}\\ 712.87\end{tabular}} &
  \begin{tabular}[c]{@{}l@{}}{\color{DRed} 3.05}\\  {\color{DGreen}109.22}\end{tabular} \\ \hline
\end{tabular}}
\vspace{-0.3 cm}
\end{table*}

\color{black}
\subsection{Model Architecture}
\label{sec:algo_model}

Consider the schematic representation of the proposed low-light image enhancement approach, \textit{SelfEnNet}, shown in Fig.~\ref{fig:blockdia}. It contains the enhancement module $\mathcal{F_E}$ and the noise-handling module $\mathcal{F_D}$. 
Fig.~\ref{fig: main_model} shows the deep convolutional neural network based model architecture of \textit{SelfEnNet}. 

\subsubsection{Enhancement module $\mathcal{F_E}$}

The model architecture of the enhancement module $\mathcal{F_E}$, as shown in Fig.~\ref{fig: main_model}, consists of two inputs and one output convolutional layers, and $4$ consecutive Residual Channel Attention Block (RCAB)~\cite{zhang2018image, woo2018cbam} blocks in between them. 
All the convolutional (`Conv3x3') layers consist of $3\times 3$ convolutions with stride $1$. Except for the output `Conv3x3' layer, all `Conv3x3' layers output $32$ channels each. The output `Conv3x3' layer generates a one-channel enhancement map from its $32$ input feature channels. All the non-linear activation functions used in $\mathcal{F_E}$ are indicated in the figure. 
As evident from the figure, RCAB is the main building block of our $\mathcal{F_E}$ model architecture. As can be seen, this block consists of two consecutive convolutional layers and one channel attention (CA) layer with a residual skip connection. The operations in this block are performed such that the channel dimension $M\times N$ is maintained throughout. It extracts suitable features leading to the estimation of the desired output at the end of the forward pass of the entire model. Fig.~\ref{fig: main_model} also shows the architecture of the CA layer used in our RCAB.

\paragraph{Residual Channel Attention Block (RCAB)}

RCAB consists of two convolution layers followed by ReLU non-linearity and a channel attention module. The input features are added to the output to constitute the residual connection. The Channel Attention (CA) module, as shown in Fig.~\ref{fig: main_model}, consists of a Global Average Pool layer, which calculates the average value of feature intensity in each channel. After that, two $1\times 1$ convolution layers (`Conv1x1') process those features, and the final sigmoid activation layer maps the importance of each channel in a range between $(0, 1)$. 
\color{black}
Then, the importance map of each channel is multiplied with input channel features to give more importance to those features, which give better optimization performance.
\color{black}

\paragraph{Feedback Mechanism}

We introduce a feedback mechanism from the $\mathcal{F_E}$ model output to the input to guide the enhancement network based on the initial estimation of enhancement map. 
The features extracted through a convolutional layer from the model output, which is enhancement map, are fed back to concatenate with the features from the first input convolutional layer. The feedback is given only once after an initial stage. In the initial stage, an enhancement map with all its location taking the value $\frac{1}{2.2}$ is considered as the model output to compute the feedback features. The sequential computation due to the feedback influenced by an intermediate output allows the model to be consistent during the iterations, resulting in a map estimate that signifies the right amount of enhancement. Note that, the value $2.2^{-1}$ is inspired from several enhancement approaches \cite{Guo2017, li2018structure, ren2020lr3m, Park2017} that use it for fixed image transformation based enhancement. Empirical studies related to the feedback mechanism are given in Section~\ref{subsec:model_ab}.

\subsubsection{Noise-handling module $\mathcal{F_D}$}

The model architecture of the noise-handling module $\mathcal{F_D}$ is shown in Fig.~\ref{fig: main_model} as well. The architecture differs from that of $\mathcal{F_E}$ in two aspects. First, there is no feedback in $\mathcal{F_D}$ from the model output, which is the three-channel denoised image. Second, the one-channel enhancement map obtained using $\mathcal{F_E}$ on the same input image as that of $\mathcal{F_D}$ is subjected to a convolution layer, and the features from the layer are concatenated with the features from the first input convolutional layer of $\mathcal{F_D}$. The use of the estimated enhancement map allows $\mathcal{F_D}$ to work specifically on the input image regions having low-light specific noise. Rest of the architecture of $\mathcal{F_D}$ is similar to $\mathcal{F_E}$. 
All the convolutional (`Conv3x3') layers consist of $3\times 3$ convolutions with stride $1$. Except for the output `Conv3x3' layer, all `Conv3x3' layers output $32$ channels each. The output `Conv3x3' layer generates the denoised image from its $32$ input feature channels. Similar to the enhancement module $\mathcal{F_E}$, there are four RCABs in the noise-handling module $\mathcal{F_D}$.

\paragraph{Enhancement Map Features}
\label{enh_map_feat}
As mentioned above, we pass the enhancement map information into the denoising network for guidance during the minimization of the losses for training. The enhancement map carries the information about the well-lit and low-light regions of an image. Thus, its guidance will help in emphasizing denoising at the low-light regions, as our denoising self-conditioning loss from Section~\ref{DSelfLoss} mainly constraints the network to not denoise well-lit regions. The extracted features from the enhancement map are concatenated with the extracted features from the noisy low-light image which is the input to the denoising network. Empirical studies on this use of enhancement map features are given in Section~\ref{ablation_enh_map}.

\color{black}

\begin{figure*}[h]
  \centering
   \subfloat{
    \includegraphics[width=0.095\textwidth]{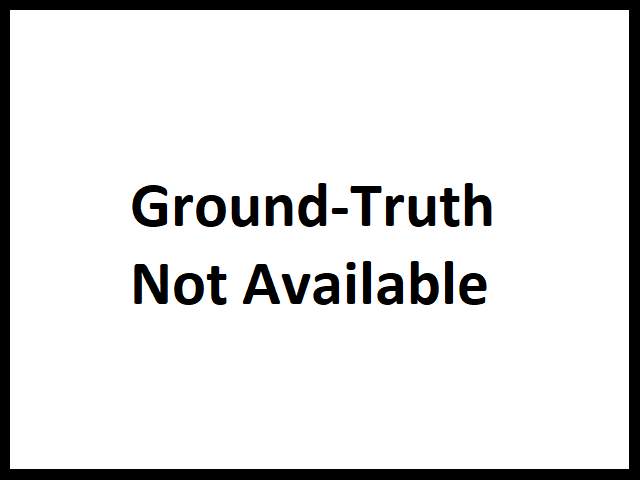}
  }\hspace{-0.3cm}
  \subfloat{
    \includegraphics[width=0.095\textwidth]{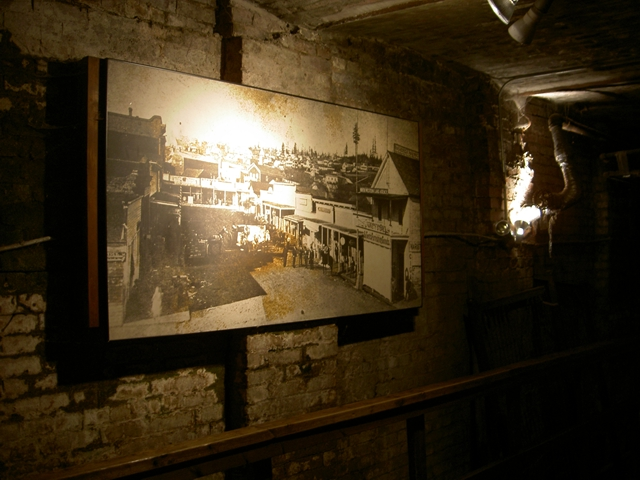}
  }\hspace{-0.3cm}
  \subfloat{
    \includegraphics[width=0.095\textwidth]{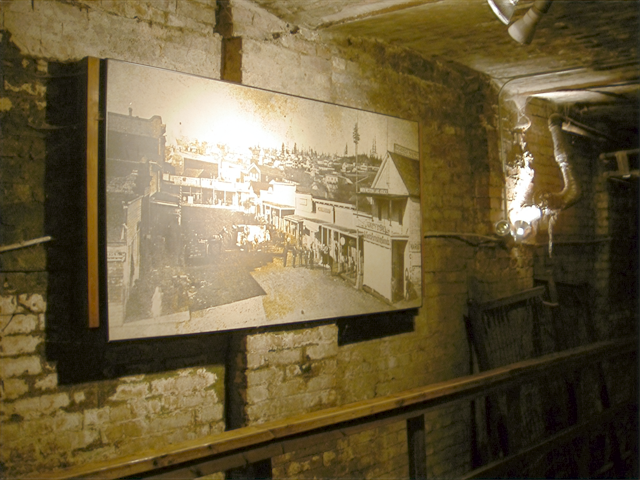}
  }\hspace{-0.3cm}
  \subfloat{
    \includegraphics[width=0.095\textwidth]{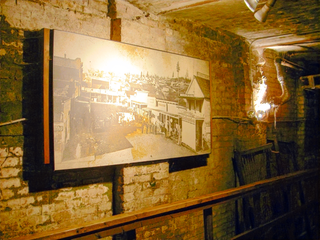}
  }\hspace{-0.3cm}
  \subfloat{
    \includegraphics[width=0.095\textwidth]{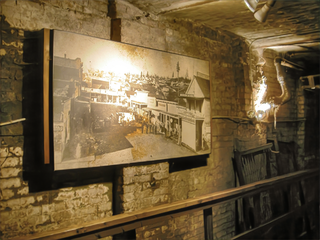}
  }\hspace{-0.3cm}
  \subfloat{
    \includegraphics[width=0.095\textwidth]{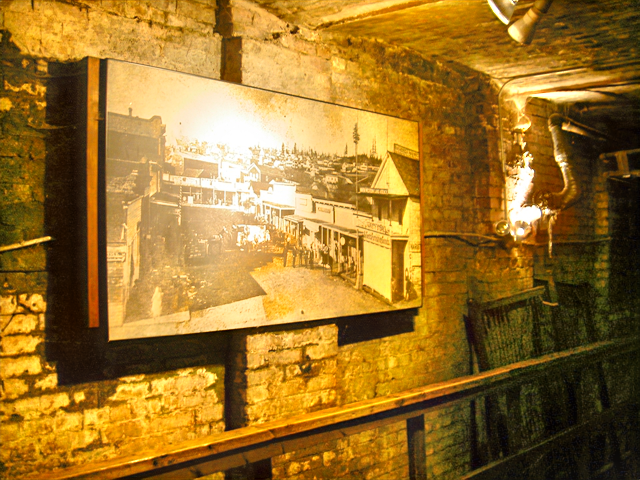}
  }\hspace{-0.3cm}
   \subfloat{
    \includegraphics[width=0.095\textwidth]{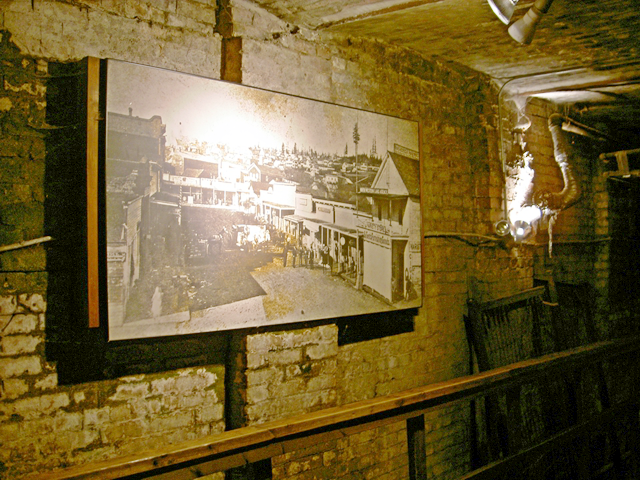}
  }\hspace{-0.3cm}
   \subfloat{
    \includegraphics[width=0.095\textwidth]{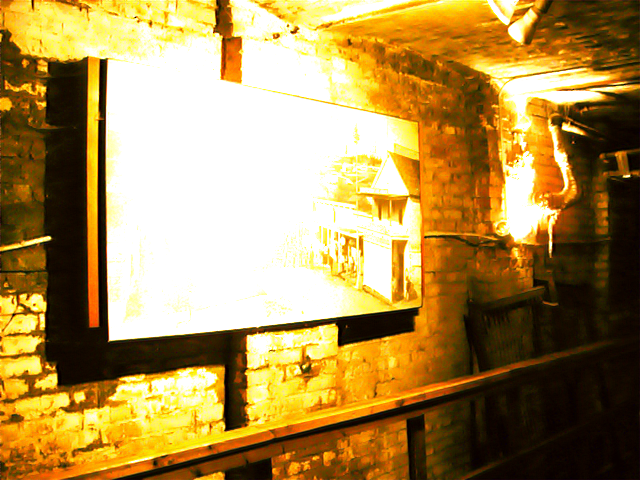}
  }\hspace{-0.3cm}
  \subfloat{
    \includegraphics[width=0.095\textwidth]{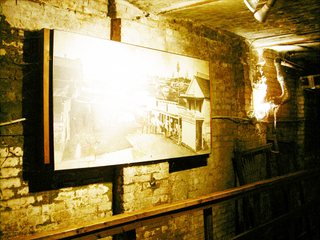}
  }\hspace{-0.3cm}
  \subfloat{
    \includegraphics[width=0.095\textwidth]{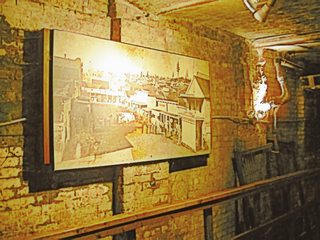}
  }\hspace{-0.3cm}
 \\\vspace{-0.2cm}
   \subfloat{
    \includegraphics[width=0.095\textwidth]{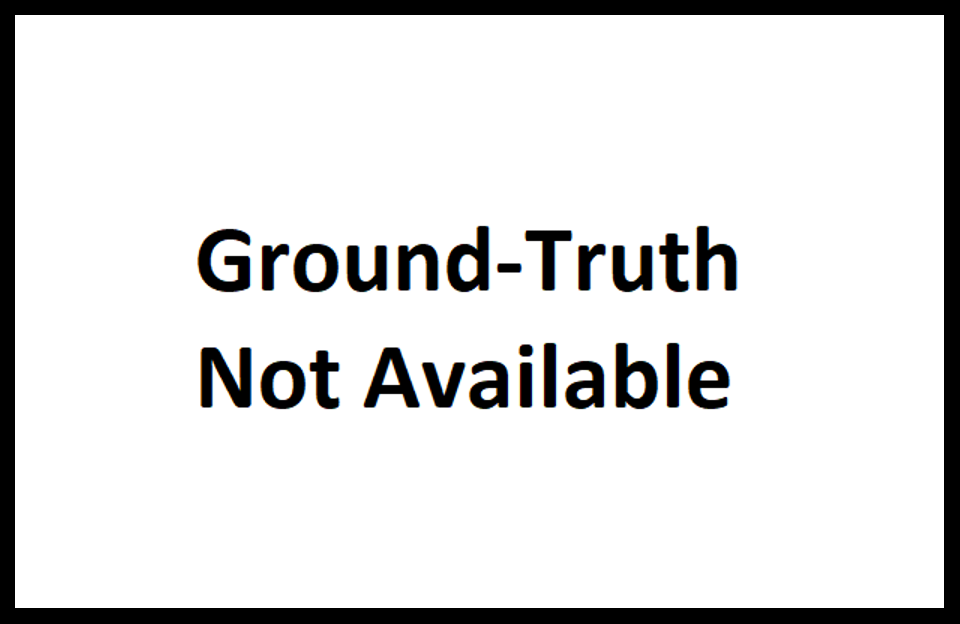}
  }\hspace{-0.3cm}
  \subfloat{
    \includegraphics[width=0.095\textwidth]{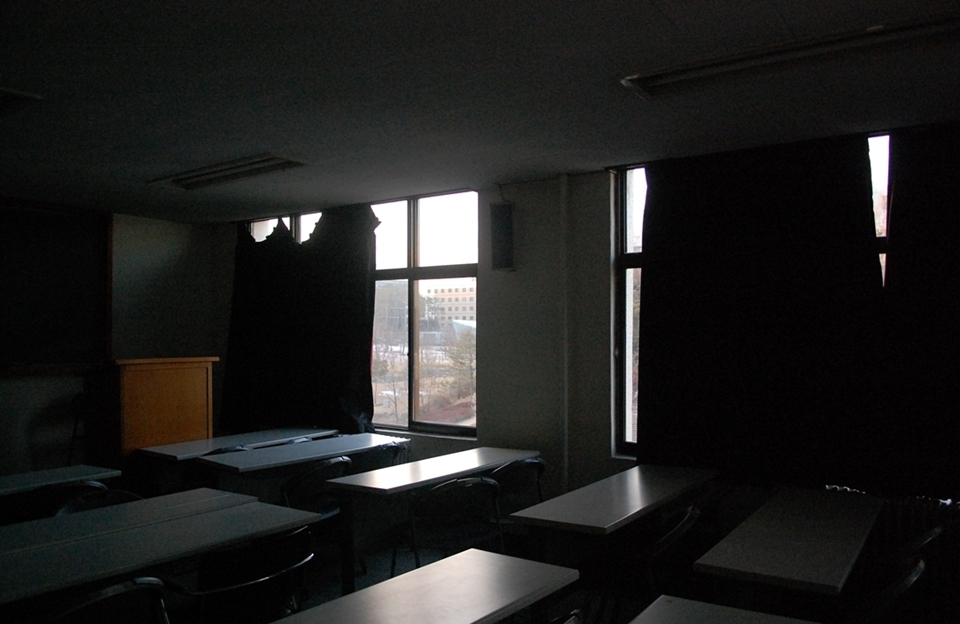}
  }\hspace{-0.3cm}
  \subfloat{
    \includegraphics[width=0.095\textwidth]{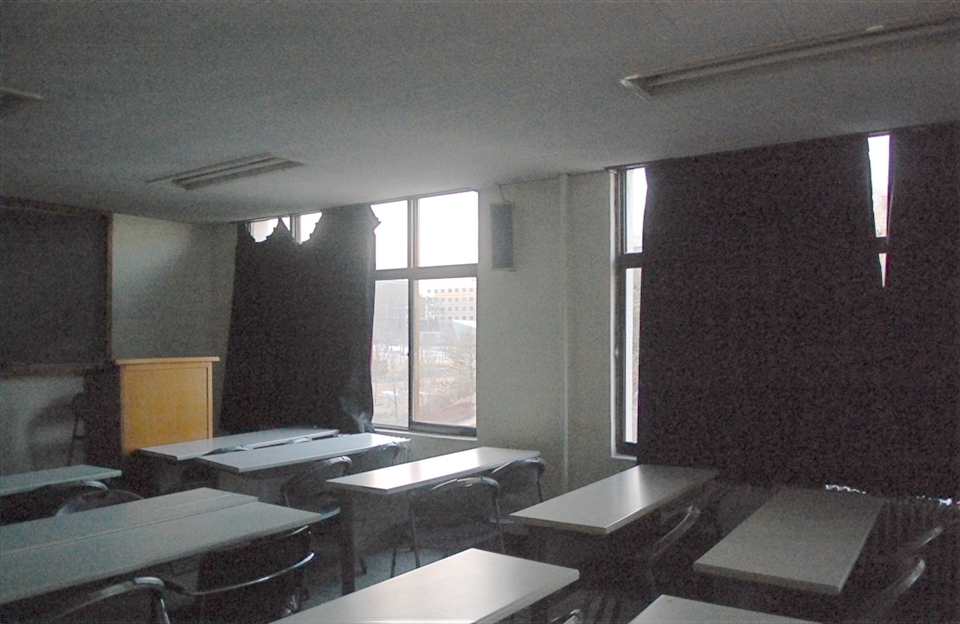}
  }\hspace{-0.3cm}
  \subfloat{
    \includegraphics[width=0.095\textwidth]{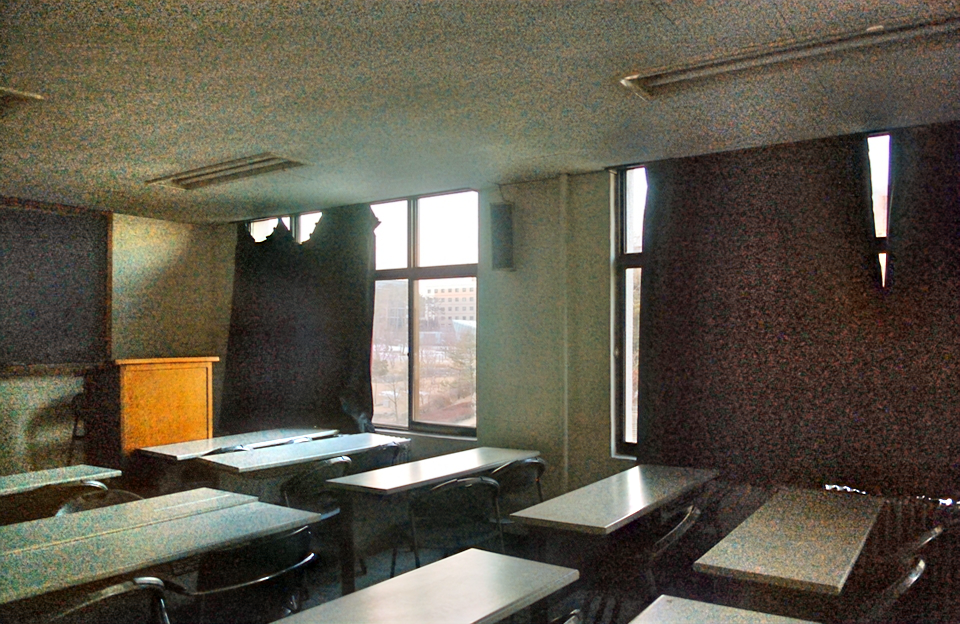}
  }\hspace{-0.3cm}
  \subfloat{
    \includegraphics[width=0.095\textwidth]{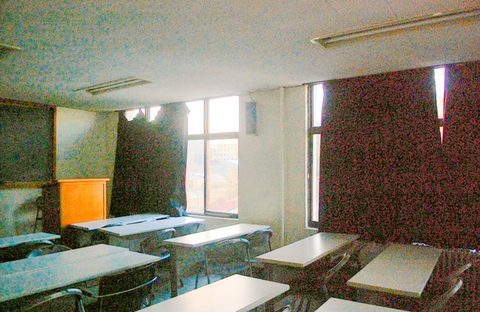}
  }\hspace{-0.3cm}
  \subfloat{
    \includegraphics[width=0.095\textwidth]{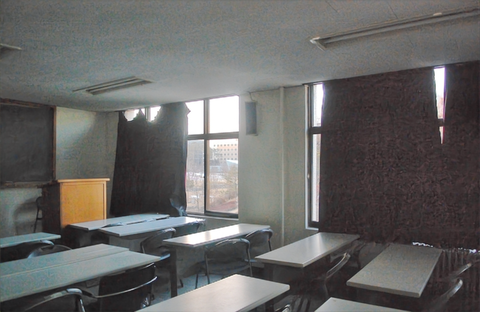}
  }\hspace{-0.3cm}
   \subfloat{
    \includegraphics[width=0.095\textwidth]{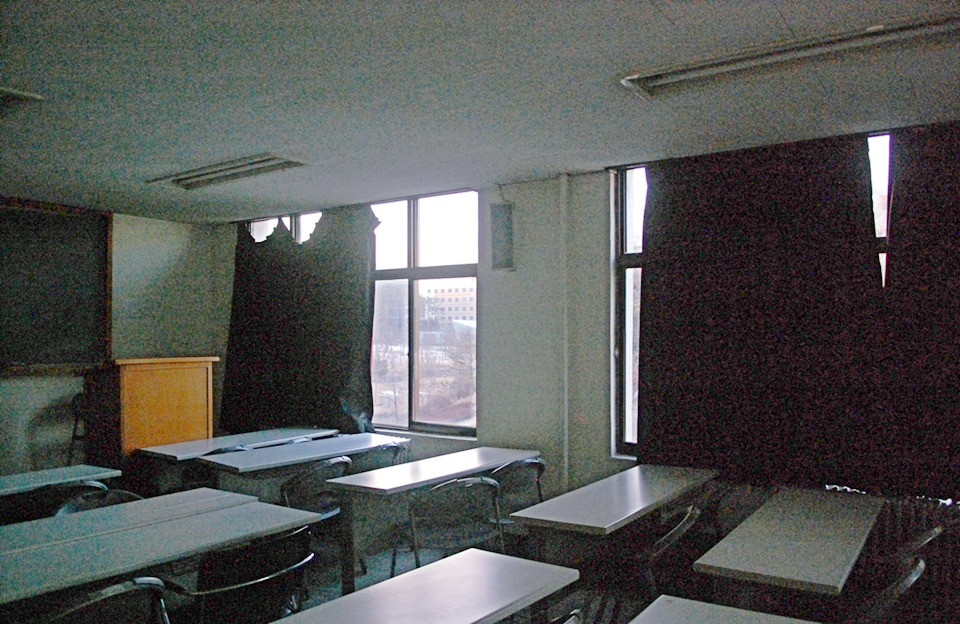}
  }\hspace{-0.3cm}
   \subfloat{
    \includegraphics[width=0.095\textwidth]{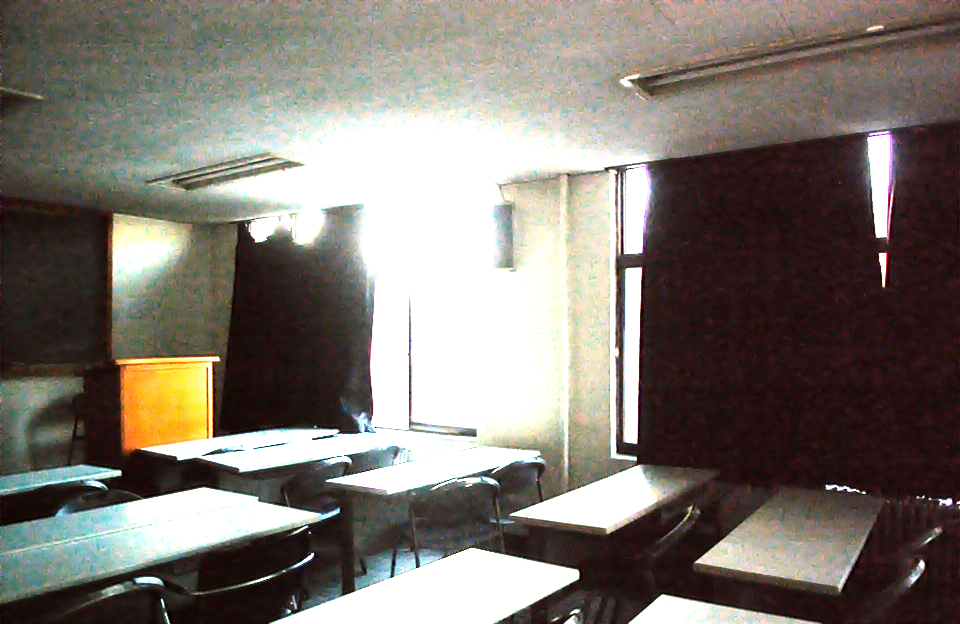}
  }\hspace{-0.3cm}
  \subfloat{
    \includegraphics[width=0.095\textwidth]{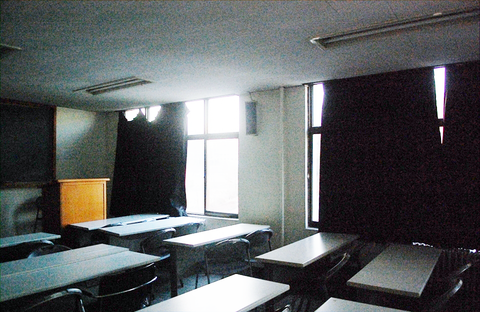}
  }\hspace{-0.3cm}
  \subfloat{
    \includegraphics[width=0.095\textwidth]{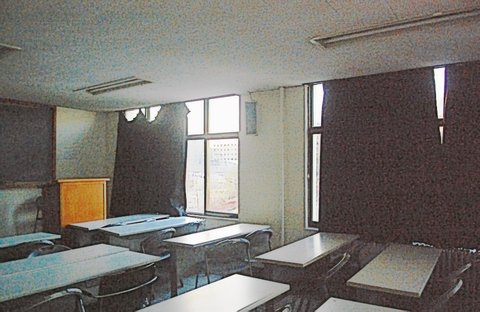}
  }\hspace{-0.3cm}
  \\\vspace{-0.2cm}
  \subfloat{
    \includegraphics[width=0.095\textwidth]{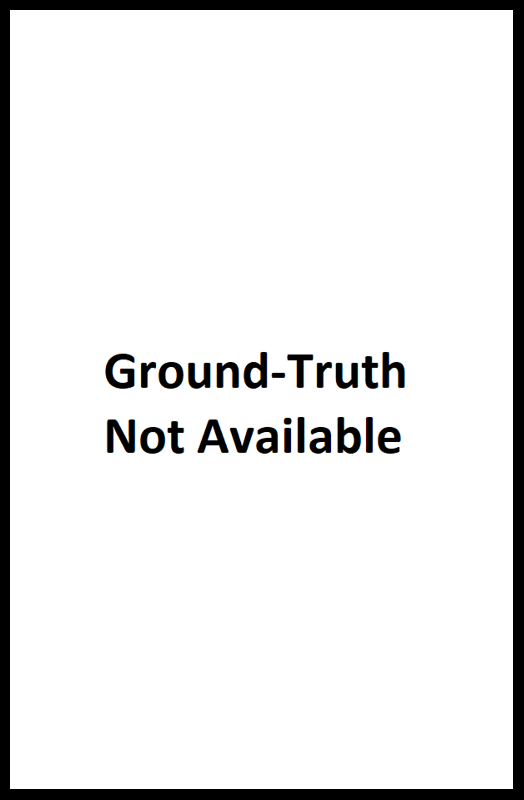}
  }\hspace{-0.3cm}
  \subfloat{
    \includegraphics[width=0.095\textwidth]{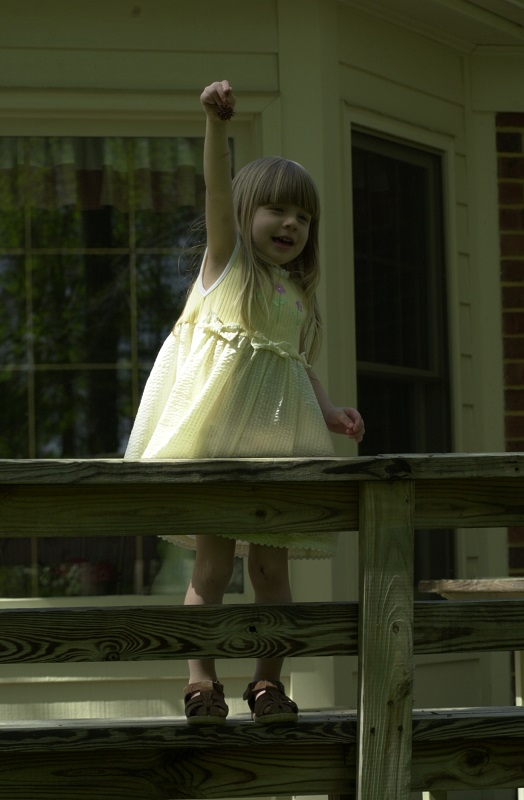}
  }\hspace{-0.3cm}
  \subfloat{
    \includegraphics[width=0.095\textwidth]{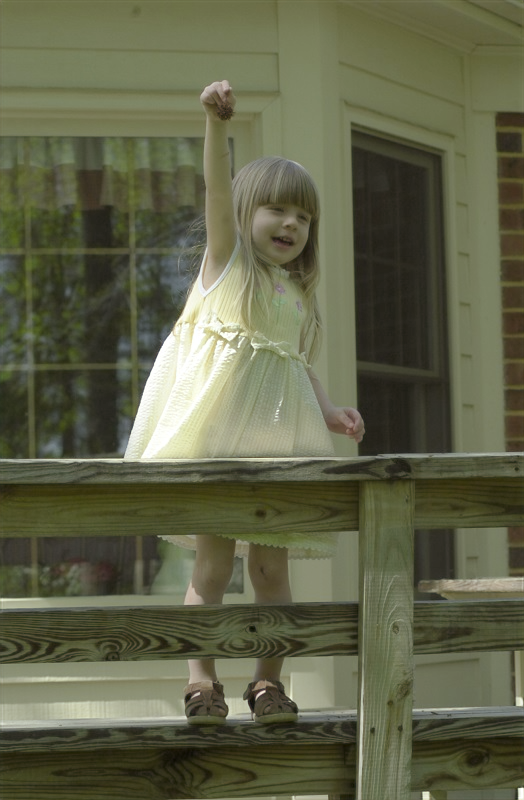}
  }\hspace{-0.3cm}
  \subfloat{
    \includegraphics[width=0.095\textwidth]{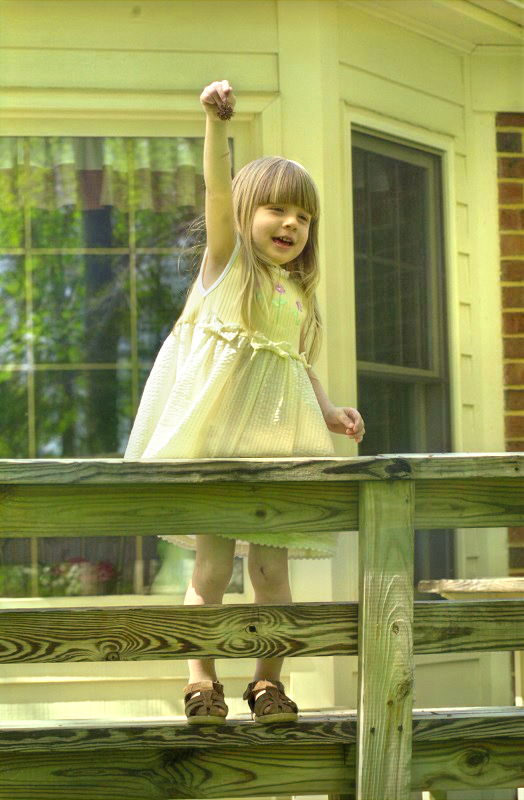}
  }\hspace{-0.3cm}
  \subfloat{
    \includegraphics[width=0.095\textwidth]{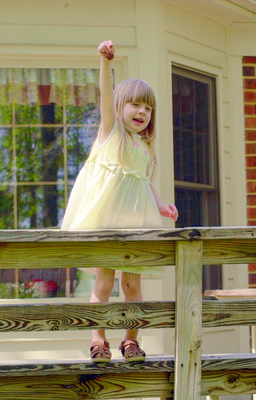}
  }\hspace{-0.3cm}
  \subfloat{
    \includegraphics[width=0.095\textwidth]{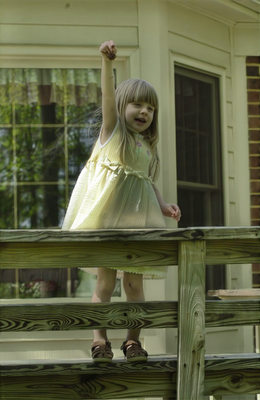}
  }\hspace{-0.3cm}
   \subfloat{
    \includegraphics[width=0.095\textwidth]{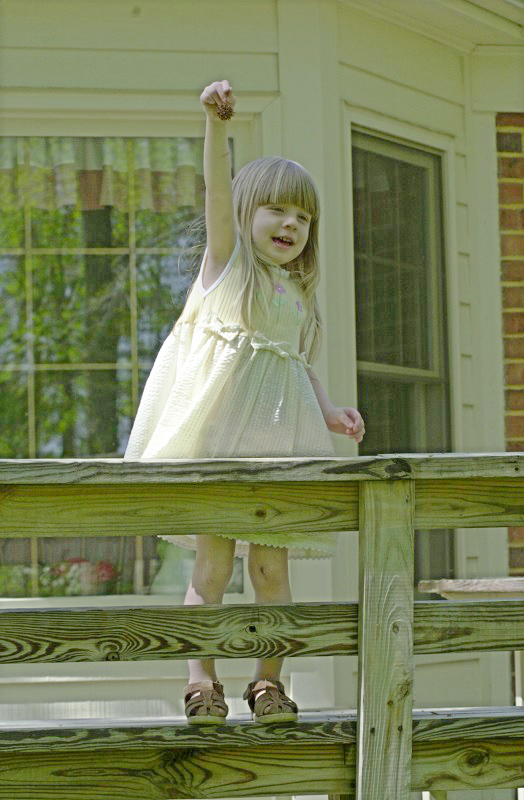}
  }\hspace{-0.3cm}
   \subfloat{
    \includegraphics[width=0.095\textwidth]{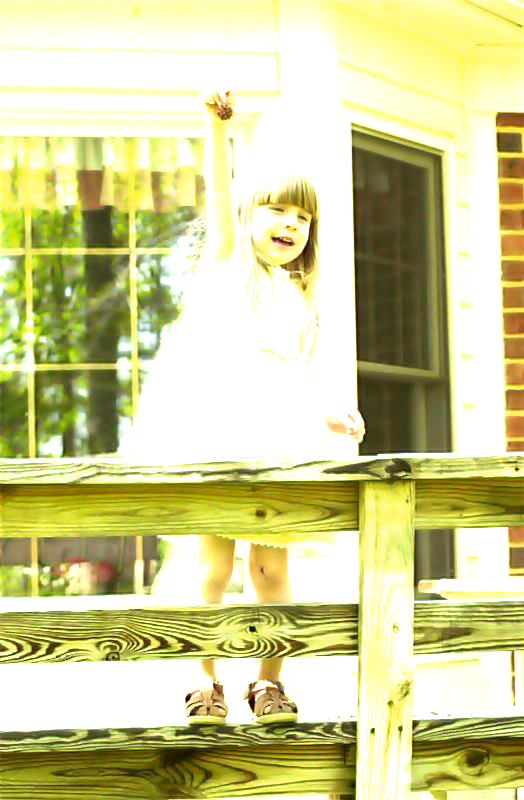}
  }\hspace{-0.3cm}
  \subfloat{
    \includegraphics[width=0.095\textwidth]{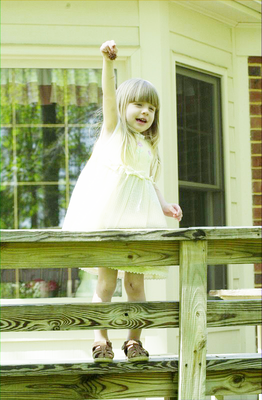}
  }\hspace{-0.3cm}
  \subfloat{
    \includegraphics[width=0.095\textwidth]{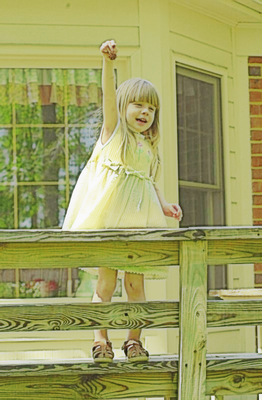}
  }\hspace{-0.3cm}
  \\\vspace{-0.2cm}
  \subfloat{
    \includegraphics[width=0.095\textwidth]{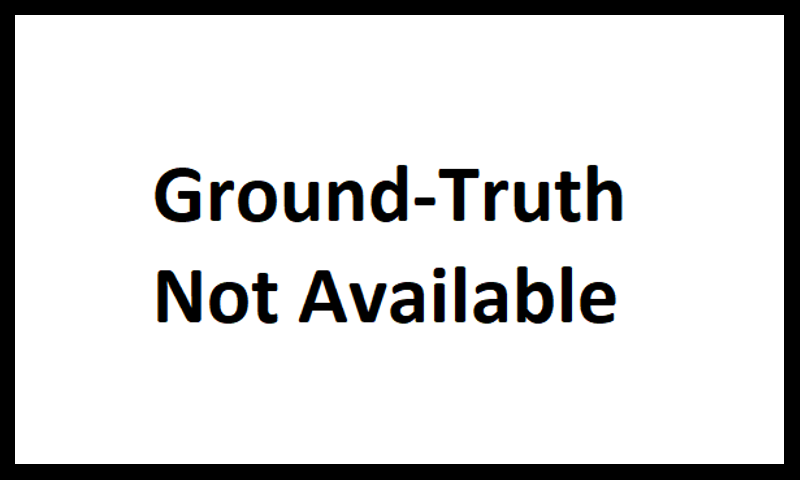}
  }\hspace{-0.3cm}
  \subfloat{
    \includegraphics[width=0.095\textwidth]{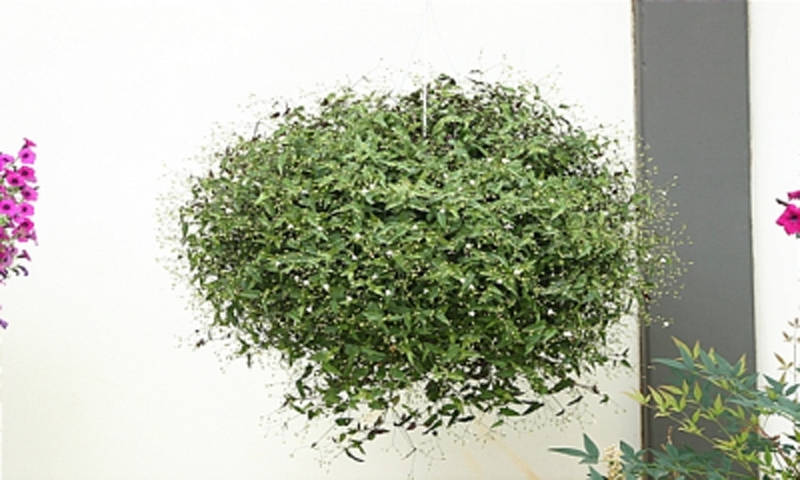}
  }\hspace{-0.3cm}
  \subfloat{
    \includegraphics[width=0.095\textwidth]{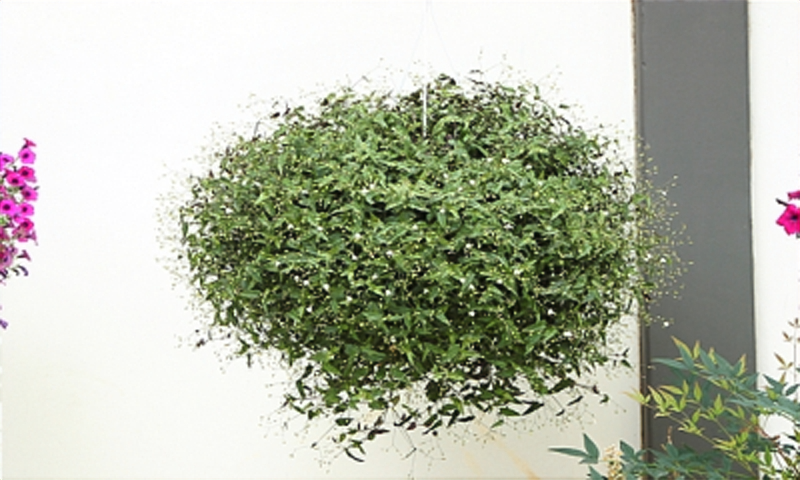}
  }\hspace{-0.3cm}
  \subfloat{
    \includegraphics[width=0.095\textwidth]{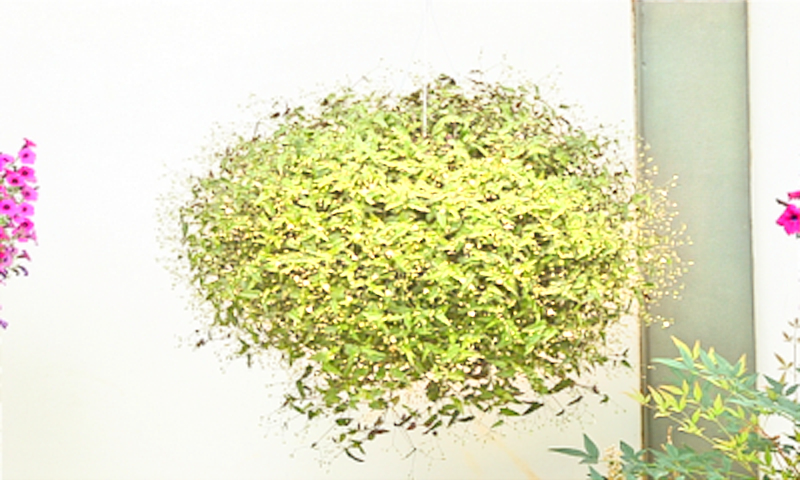}
  }\hspace{-0.3cm}
  \subfloat{
    \includegraphics[width=0.095\textwidth]{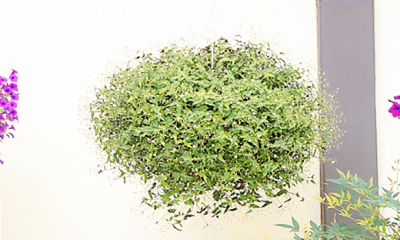}
  }\hspace{-0.3cm}
  \subfloat{
    \includegraphics[width=0.095\textwidth]{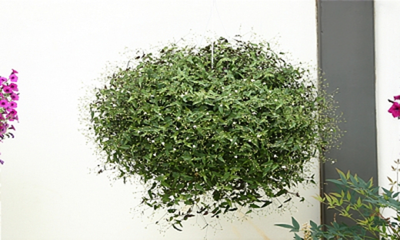}
  }\hspace{-0.3cm}
   \subfloat{
    \includegraphics[width=0.095\textwidth]{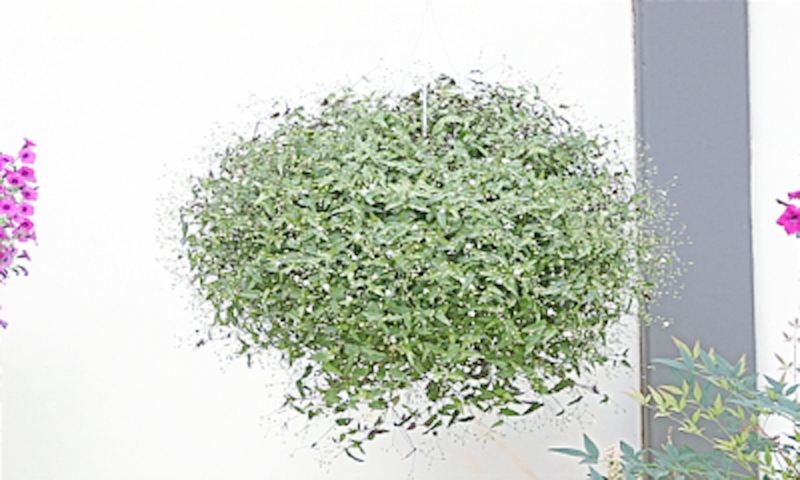}
  }\hspace{-0.3cm}
   \subfloat{
    \includegraphics[width=0.095\textwidth]{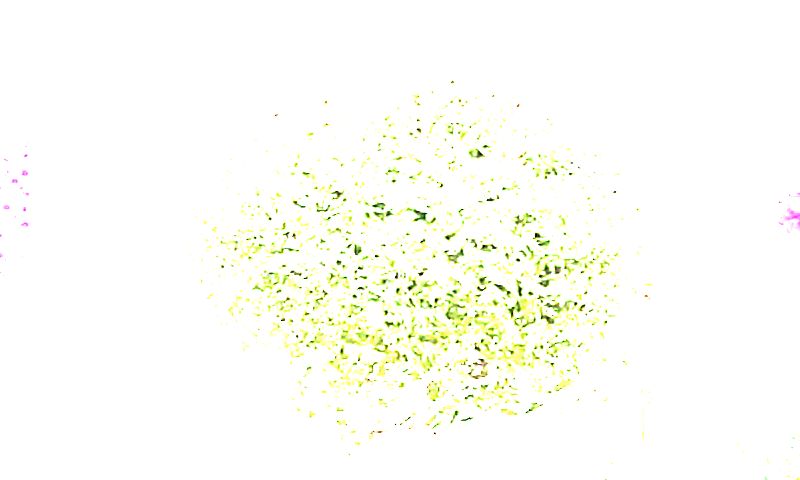}
  }\hspace{-0.3cm}
  \subfloat{
    \includegraphics[width=0.095\textwidth]{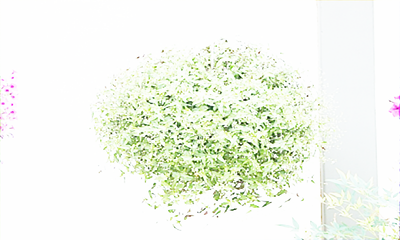}
  }\hspace{-0.3cm}
  \subfloat{
    \includegraphics[width=0.095\textwidth]{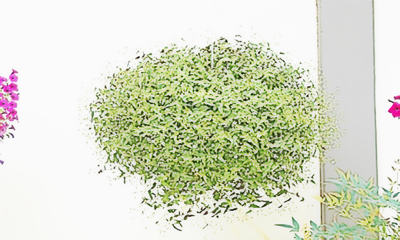}
  }\hspace{-0.3cm}
  \\\vspace{-0.2cm}
  \subfloat{
    \includegraphics[width=0.095\textwidth]{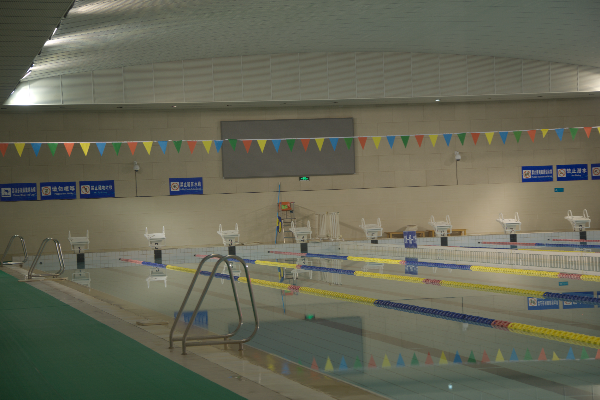}
  }\hspace{-0.3cm}
  \subfloat{
    \includegraphics[width=0.095\textwidth]{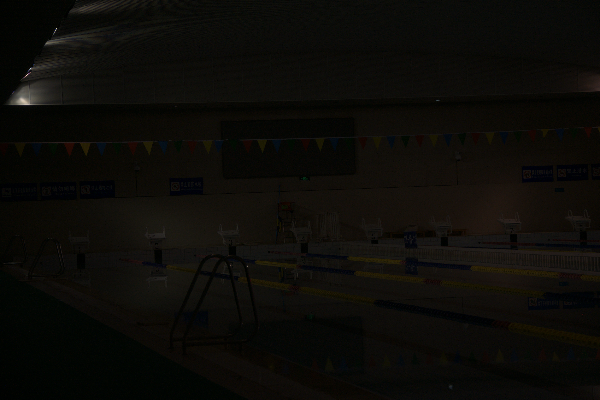}
  }\hspace{-0.3cm}
  \subfloat{
    \includegraphics[width=0.095\textwidth]{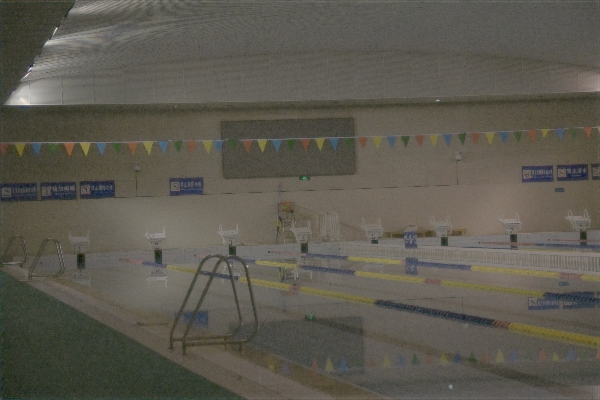}
  }\hspace{-0.3cm}
  \subfloat{
    \includegraphics[width=0.095\textwidth]{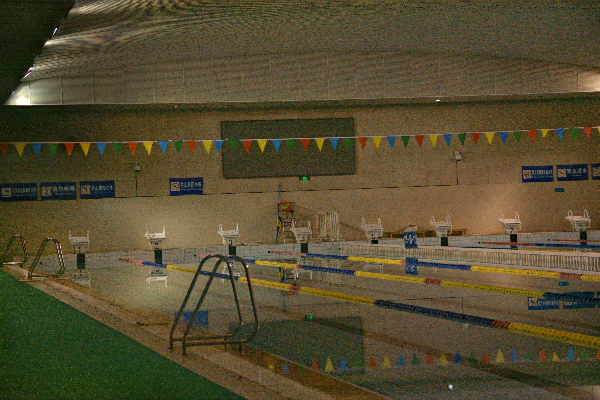}
  }\hspace{-0.3cm}
  \subfloat{
    \includegraphics[width=0.095\textwidth]{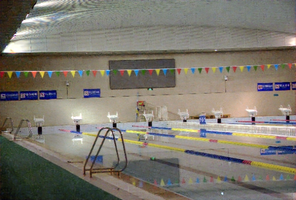}
  }\hspace{-0.3cm}
  \subfloat{
    \includegraphics[width=0.095\textwidth]{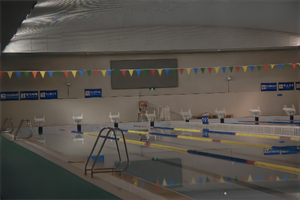}
  }\hspace{-0.3cm}
   \subfloat{
    \includegraphics[width=0.095\textwidth]{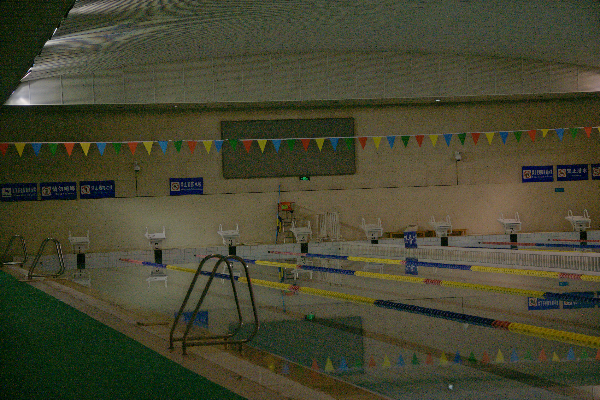}
  }\hspace{-0.3cm}
   \subfloat{
    \includegraphics[width=0.095\textwidth]{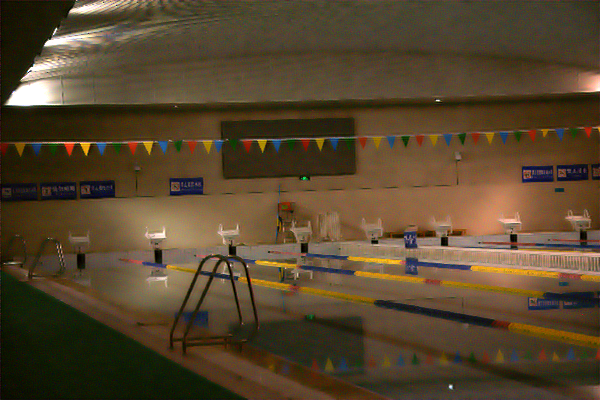}
  }\hspace{-0.3cm}
  \subfloat{
    \includegraphics[width=0.095\textwidth]{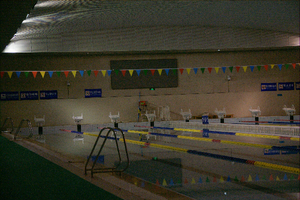}
  }\hspace{-0.3cm}
  \subfloat{
    \includegraphics[width=0.095\textwidth]{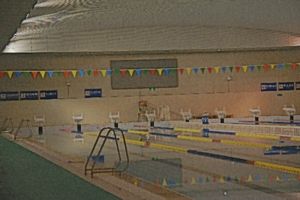}
  }\hspace{-0.3cm}
 \\\vspace{-0.2cm}
   \subfloat{
    \includegraphics[width=0.095\textwidth]{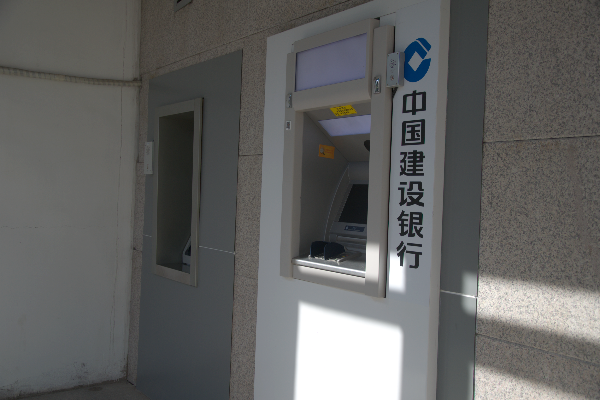}
  }\hspace{-0.3cm}
  \subfloat{
    \includegraphics[width=0.095\textwidth]{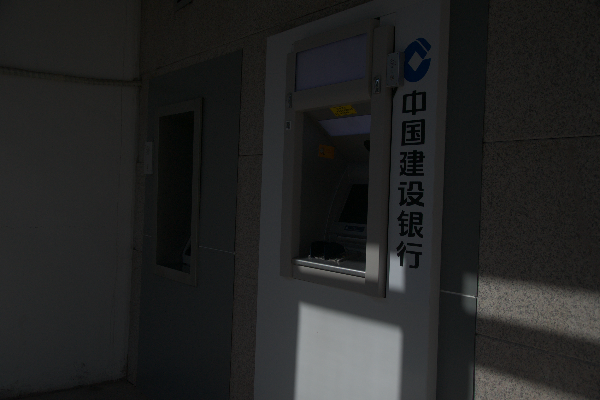}
  }\hspace{-0.3cm}
  \subfloat{
    \includegraphics[width=0.095\textwidth]{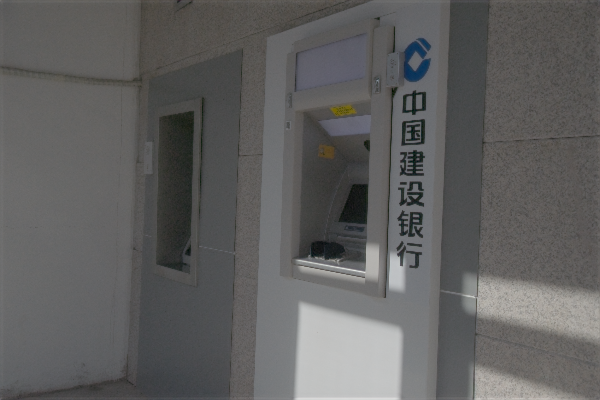}
  }\hspace{-0.3cm}
  \subfloat{
    \includegraphics[width=0.095\textwidth]{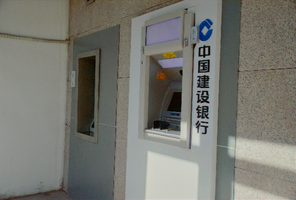}
  }\hspace{-0.3cm}
  \subfloat{
    \includegraphics[width=0.095\textwidth]{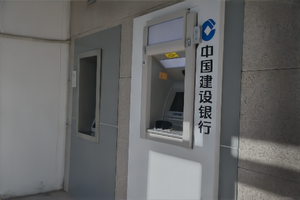}
  }\hspace{-0.3cm}
  \subfloat{
    \includegraphics[width=0.095\textwidth]{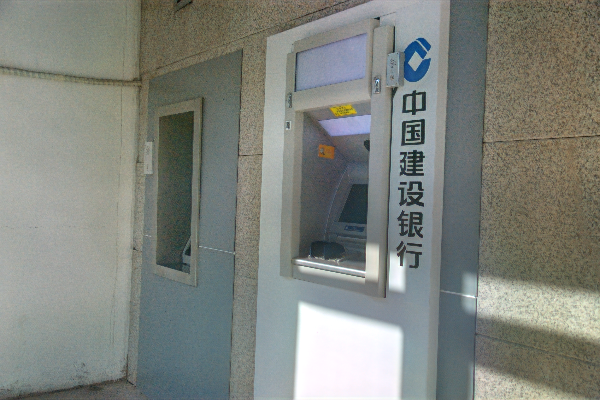}
  }\hspace{-0.3cm}
   \subfloat{
    \includegraphics[width=0.095\textwidth]{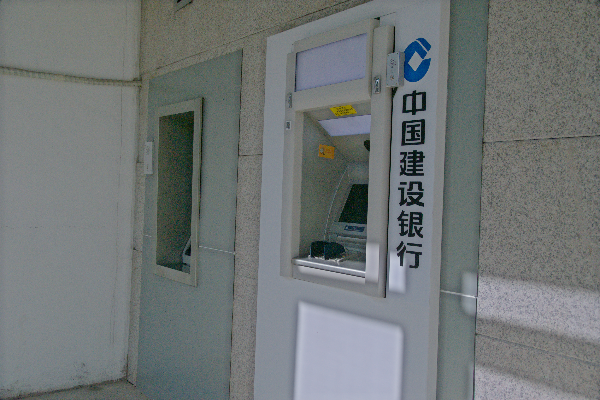}
  }\hspace{-0.3cm}
   \subfloat{
    \includegraphics[width=0.095\textwidth]{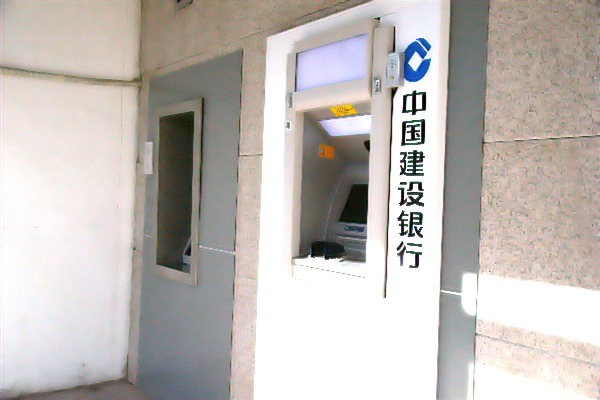}
  }\hspace{-0.3cm}
  \subfloat{
    \includegraphics[width=0.095\textwidth]{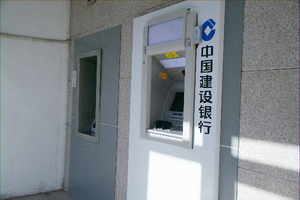}
  }\hspace{-0.3cm}
  \subfloat{
    \includegraphics[width=0.095\textwidth]{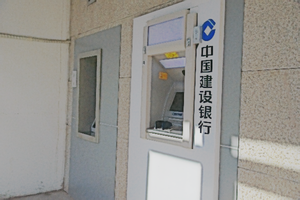}
  }\hspace{-0.3cm}
  \\\vspace{-0.2cm}
   {\setcounter{subfigure}{0}}
   \subfloat[ GT]{
     \includegraphics[width=0.095\textwidth]{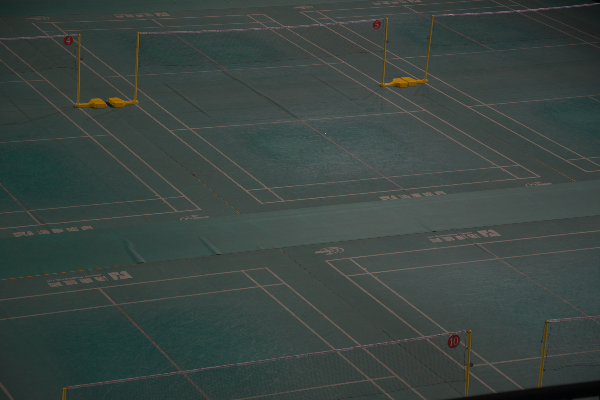}
   }\hspace{-0.3cm}
   \subfloat[ Input]{
     \includegraphics[width=0.095\textwidth]{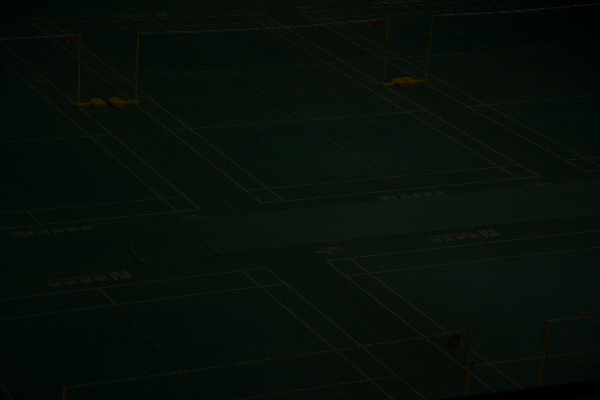}
   }\hspace{-0.3cm}
   \subfloat[Ours$^\S$]{
     \includegraphics[width=0.095\textwidth]{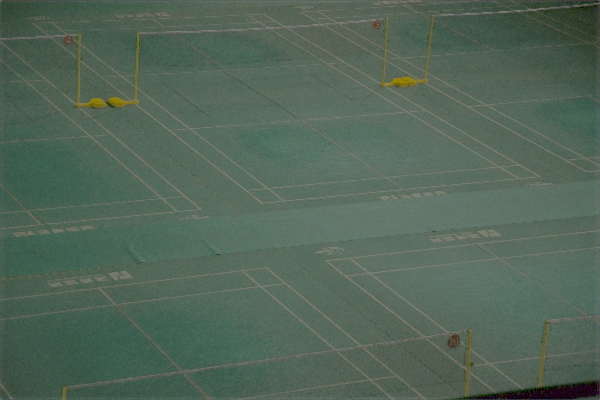}
   }\hspace{-0.3cm}
   \subfloat[MLLEN$^*$]{
     \includegraphics[width=0.095\textwidth]{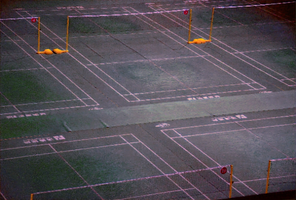}
   }\hspace{-0.3cm}
   \subfloat[Bread$^*$]{
     \includegraphics[width=0.095\textwidth]{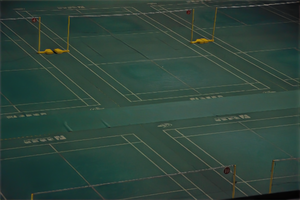}
   }\hspace{-0.3cm}
   \subfloat[EGAN$^\S$]{
     \includegraphics[width=0.095\textwidth]{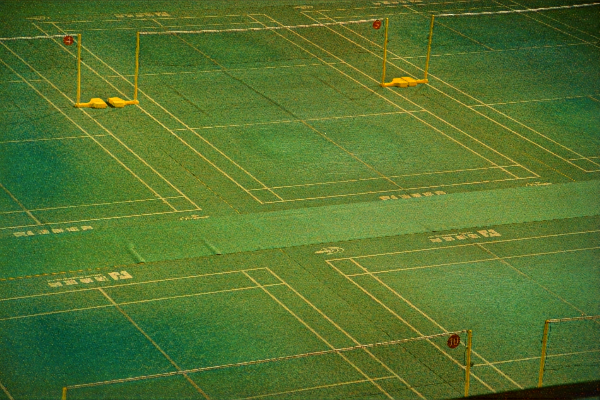}
   }\hspace{-0.3cm}
   \subfloat[ZeroDCE$^+$]{
     \includegraphics[width=0.095\textwidth]{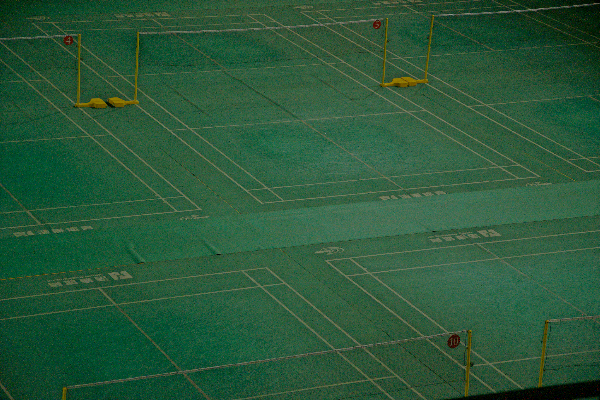}
   }\hspace{-0.3cm}
   \subfloat[RUAS$^+$]{
     \includegraphics[width=0.095\textwidth]{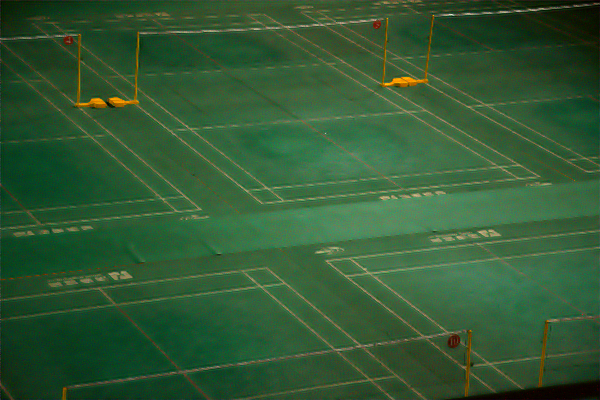}
   }\hspace{-0.3cm}
   \subfloat[SCI$^+$]{
     \includegraphics[width=0.095\textwidth]{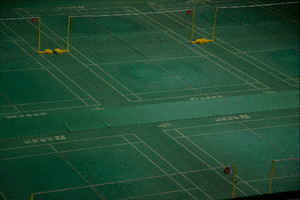}
   }\hspace{-0.3cm}
   \subfloat[PairLIE$^+$]{
     \includegraphics[width=0.095\textwidth]{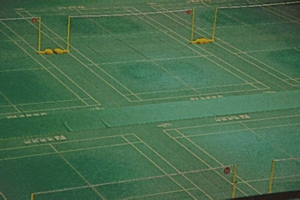}
   }
  \caption{Enhancement of (b) low-light images from different datasets and their corresponding (a) ground truths. $*$ denotes a technique with paired supervision, $+$ denotes a  technique without paired and unpaired supervision and $\S$ denotes a technique only with unpaired supervision. \color{black}The first four images are real-world low-light images whose ground truths are not available.}
  \label{fig:SubjGT} 
\end{figure*}


\section{Experimental Results and Discussion}
\label{sec:exp_results}
In this section, we compare our proposed network for low-light image enhancement, \textit{SelfEnNet}, to the state-of-the-art approaches of DRD~\cite{wei2018deep}, DRBN~\cite{yang2021band}, URN~\cite{wu2022uretinex}, MLLEN~\cite{fan2022multiscale} and Bread~\cite{guo2023low} that require paired supervision, EGAN~\cite{jiang2021enlightengan} that uses unpaired supervision, and LIME~\cite{Guo2017}, RRM~\cite{li2018structure}, ALSM~\cite{wang2019low}, Zero-DCE~\cite{guo2020zero}, ZCP~\cite{kar2021zero}, RUAS~\cite{liu2021retinex}, SCI~\cite{ma2022toward}, PairLIE~\cite{fu2023learning} which do not require paired or unpaired supervision, on multiple standard real-world image datasets. The codes provided by the respective authors are used for the comparison.

\color{black}
\subsection{Experimental Settings and Implementation Details}

\subsubsection{Dataset Description}
\label{sec:dataset}

\paragraph{The LOL real-world image dataset}

Following several state-of-the-art deep learning based low-light image enhancement approaches, we train our model using the $689$ training images from the latest LOL real-world image dataset~\cite{yang2021sparse} by using them in an unpaired manner as explained in Section~\ref{sec:algo}. We use the $345$ well-lit images and the $344$ low-light images in the set, but we do not use them for paired supervision unlike~\cite{yang2021sparse, yang2020fidelity, yang2021band}. Note that this trained model of ours is used for all the experiments in this paper, unless mentioned otherwise. We evaluate our low-light image enhancement model on the $100$ test image pairs available in the LOL dataset. As pointed out in~\cite{yang2020fidelity}, the test image pairs are from a ``real captured dataset including highly degraded images", and as~\cite{ma2021learning} suggests, the ``hardest part is that noises exist" in the test images. All the existing approaches (that require training), which are compared to our model on the $100$ test images of the LOL dataset, are trained on the training images of the LOL dataset~\cite{yang2021sparse, wei2018deep}.

\paragraph{Other standard low-light image datasets}
\label{RealDataset}

We consider $5$ other standard real-world low-light image datasets in which the images are not accompanied with corresponding references. They are the  VV\footnote{https://sites.google.com/site/vonikakis/datasets}, LIME~\cite{Guo2017}, NPE~\cite{wang2013naturalness},  Fusion~\cite{Fu2016} and DICM~\cite{WinNT} datasets with several low-light images of different varieties in them. We use the pre-trained models of all the approaches as provided by the respective authors for comparing them to our model, which is pre-trained on the LOL dataset.
\begin{table*}
\centering
\caption{\color{black} Computational efficiency analysis of our SelfEnNet and comparison with the state-of-the-art deep learning models.}
\label{tab: compute}
\resizebox{\textwidth}{!}{%
\begin{tabular}{|c|ccccc|cccccc|}
\hline
\multirow{3}{*}{Measures} &
  \multicolumn{5}{c|}{\multirow{2}{*}{Paired Supervision}} &
  \multicolumn{6}{c|}{No Paired Supervision} \\ \cline{7-12} 
 &
  \multicolumn{5}{c|}{} &
  \multicolumn{4}{c|}{No Unpaired Supervision} &
  \multicolumn{2}{c|}{Unpaired Supervision} \\ \cline{2-12} 
 &
  \multicolumn{1}{c|}{\begin{tabular}[c]{@{}l@{}}DRD~\cite{wei2018deep}\\ BMVC'21\end{tabular}} &
  \multicolumn{1}{c|}{\begin{tabular}[c]{@{}l@{}}DRBN~\cite{yang2021band}\\ TIP'21\end{tabular}} &
  \multicolumn{1}{c|}{\begin{tabular}[c]{@{}l@{}}URN~\cite{wu2022uretinex}\\ CVPR'22\end{tabular}} &
  \multicolumn{1}{c|}{\begin{tabular}[c]{@{}l@{}}MLLEN~\cite{fan2022multiscale}\\ TCSVT'22\end{tabular}} &
  \begin{tabular}[c]{@{}l@{}}Bread~\cite{guo2023low}\\ IJCV'23\end{tabular} &
  \multicolumn{1}{c|}{\begin{tabular}[c]{@{}l@{}}Zero-DCE~\cite{guo2020zero}\\ CVPR'20\end{tabular}} &
  \multicolumn{1}{c|}{\begin{tabular}[c]{@{}l@{}}RUAS~\cite{liu2021retinex}\\ CVPR'21\end{tabular}} &
  \multicolumn{1}{c|}{\begin{tabular}[c]{@{}l@{}}SCI~\cite{ma2022toward}\\ CVPR'22\end{tabular}} &
  \multicolumn{1}{c|}{\begin{tabular}[c]{@{}l@{}}PairLIE~\cite{fu2023learning}\\ CVPR'23\end{tabular}} &
  \multicolumn{1}{c|}{\begin{tabular}[c]{@{}l@{}}EGAN~\cite{jiang2021enlightengan}\\ TIP'21\end{tabular}} &
  \begin{tabular}[c]{@{}l@{}}SelfEnNet\\ (Ours)\end{tabular} \\ \hline
\begin{tabular}[c]{@{}c@{}}Parameters in Millions\end{tabular} &
  \multicolumn{1}{c|}{0.56} &
  \multicolumn{1}{c|}{0.58} &
  \multicolumn{1}{c|}{0.34} &
  \multicolumn{1}{c|}{12.15} &
  2.12 &
  \multicolumn{1}{c|}{0.079} &
  \multicolumn{1}{c|}{0.003} &
  \multicolumn{1}{c|}{0.0003} &
  \multicolumn{1}{c|}{0.34} &
  \multicolumn{1}{c|}{8.64} &
  0.17 \\ \hline
\begin{tabular}[c]{@{}c@{}}Memory in MB\end{tabular} &
  \multicolumn{1}{c|}{214.56} &
  \multicolumn{1}{c|}{67.98} &
  \multicolumn{1}{c|}{134.31} &
  \multicolumn{1}{c|}{251.44} &
  72.35 &
  \multicolumn{1}{c|}{78.73} &
  \multicolumn{1}{c|}{30.96} &
  \multicolumn{1}{c|}{4.51} &
  \multicolumn{1}{c|}{37.45} &
  \multicolumn{1}{c|}{132.20} &
  43.60 \\ \hline
\begin{tabular}[c]{@{}c@{}}Inference Time in ms\end{tabular} &
  \multicolumn{1}{c|}{11.91} &
  \multicolumn{1}{c|}{21.76} &
  \multicolumn{1}{c|}{40.02} &
  \multicolumn{1}{c|}{22.93} &
  17.63 &
  \multicolumn{1}{c|}{3.11} &
  \multicolumn{1}{c|}{5.47} &
  \multicolumn{1}{c|}{1.52} &
  \multicolumn{1}{c|}{8.32} &
  \multicolumn{1}{c|}{5.91} &
  7.93 \\ \hline

\end{tabular}%
}
\end{table*}

\subsubsection{Model Training Details}
\label{subsec:exp_setting}

We use two different networks in our model, the enhancement network and the noise-handling network. As required, the enhancement network is trained first, followed by the noise-handling network. During the training of the denoising network, the weights of the trained enhancement network are kept frozen. The already-trained enhancement network is only used to generate the enhancement map for the denoising network, which can be seen in Fig.~\ref{fig: main_model}.
In both the cases, the network is trained for $250$ epochs and each epoch consists of $1000$ batch updates. $128\times 128$ size patches are collected in each batch update and the batch size is $4$. During the training, those patches are augmented using random horizontal and vertical flipping along with $90^{\circ}$ rotation.
The loss functions as mentioned in Section~\ref{SSloss},~\ref{LSCloss}, and~\ref{WSCloss} are used to calculate the losses for the enhancement network. The loss functions from Section~\ref{GLoss},~\ref{FLoss}, and ~\ref{DSelfLoss} are used for calculating the losses of denoising network. 
The Adam optimizer~\cite{kingma2014adam} with the default settings in the PyTorch environment is employed as the update rule for the weight parameters. The initial learning rate is set to $10^{-4}$, and learning rate is halved after every $50$ epochs. Although we extract patches during training, we feed the lowlight images as a whole into the network while performing the denoising and enhancement on the testing set of lowlight images in the dataset considered. Our model code is publicly available at \href{https://github.com/aupendu/SelfEnNet}{https://github.com/aupendu/SelfEnNet} for the reproducibility of our work.
\color{black}

\subsection{Comparative Performance Analysis}
\label{subsec:result}

\paragraph{Quantitative Evaluation}
We perform reference-based quantitative evaluation of all the techniques, including ours, on the popular LOL dataset~\cite{yang2021sparse}. Following the standard practice, we consider the widely used quality measures PSNR, SSIM~\cite{Wang2004SSIM},  CIEDE~\cite{sharma2005ciede2000} and LPIPS$_\textrm{VGG}$~\cite{zhang2018unreasonable}. Higher PSNR and SSIM are better, and for CIEDE and LPIPS$_\textrm{VGG}$ that quantify color preservation and perceptual similarity, respectively, lower values represent better performance. Table~\ref{tab: LOL} shows that our low-light image enhancement network outperforms all the techniques without paired supervision in terms of all four measures. In terms of PSNR, it performs better than all the others except just one pair-supervised approach. It also does well in terms of SSIM, CIEDE, and LPIPS$_\textrm{VGG}$, even performing better than a couple of pair-supervised approaches. The performance values of our approach indicate that it produces output images of satisfactory quality while preserving the perceptual appearance and handling the noise satisfactorily.

Now, compared to all approaches that do not need paired supervision, our approach performs closest to the best pair-supervised approaches, Bread and URN. As our approach does not need paired supervision, it can be trained using any set of low-light images unrelated to the kind of well-lit images desired. 
For the same reason, it is also expected to outperform the state-of-the-art including URN, DRBN and Bread on low-light images of different kinds than those used in their training. 
An experiment is performed in this regard, whose results are shown in Table~\ref{tab: Real}.

In Table~\ref{tab: Real}, we evaluate the enhancement techniques on the standard low-light image databases discussed in Section~\ref{sec:dataset}, which do not have corresponding well-lit references. Therefore, we use the widely used no-reference quality measures namely NIQE~\cite{Mittal2013} and LOE~\cite{wang2013naturalness}, and present the related results in the table. NIQE quantifies the image quality in terms of naturalness and LOE, which is specific to low-light image enhancement, quantifies the image quality in terms of the preservation of lightness order. Table~\ref{tab: Real} shows that our network comprehensively outperforms all the state-of-the-art approaches in terms of LOE for all the $5$ datasets. The results also show that our approach either performs the best or close to the best performing techniques (ALSM for DICM dataset, EGAN for LIME dataset) in terms of NIQE, yielding measures within the top-2 for all the $5$ datasets. The observed superiority of our approach over the pair-supervised approaches URN, DRBN and Bread signifies that the superiority is due to the robustness of our approach in mapping low-light images to well-lit ones.

\paragraph{Subjective Evaluation}
The low-light image enhancement results of all the approaches including ours on a few images from the datasets are shown in Fig.~\ref{fig:SubjGT}, from which we evaluate them subjectively. We find that the results of our approach represent sufficient enhancement of the images that reveals the details without introducing any significant artifact. From the images in the last three rows of the figure where the ground truths are available, we see that our approach does almost as good as the pair-supervised approaches in providing results close to the ground truth. Further, from the $1^{st}$ and $3^{rd}$ row images, we see that our approach does not produce results where the details (observe the attires) are diminished unlike a few other techniques. Our approach also avoids over-enhancement unlike a few others. This is evident from the $4^{th}$ row images where only a little enhancement is required and from the $5^{th}$ row images by observing the ground truth.

\color{black}
\subsection{Computational Efficiency}
\label{subsec:compute}

We perform computational efficiency analysis of our model as compared to other deep learning models. Table~\ref{tab: compute} shows the model parameters in millions, GPU memory consumption in MegaBytes (MB) during inferencing, and inference time in milliseconds (ms). The models have been implemented on an NVIDIA 2080Ti GPU with PyTorch Framework. The GPU memory requirement and inference time are calculated for an input image of size $256\times 256$. Table~\ref{tab: compute} shows that the complexity of our proposed framework is better than the state-of-the-art supervised techniques and comparable to most other `no paired supervision' models. 
\color{black}

\subsection{Ablation Studies}
\label{subsec:ablation}

\subsubsection{Importance of the Loss Functions for our Enhancement Module}
\label{sec:ELoss}
Employing the test images of the LOL dataset, Table~\ref{tab: enh_ablation} demonstrates the importance of each of the loss functions, namely, enhancement self-supervision $\mathcal{L_{SS}}$, enhancement self-conditioning $\mathcal{L_{SC}}$ and well-lit self-conditioning $\mathcal{L_{WSC}}$ losses employed in our enhancement module, whose rationale has been provided in Section~\ref{EMAPEstimation}. While Ablations 2, 3 and 4 show that the most effective loss function is $\mathcal{L_{SS}}$ when a combination of the losses are used, Ablation 1 suggests that $\mathcal{L_{SS}}$ alone may not be useful. It is evident that among the combinations of two loss functions, Ablation 4 having $\mathcal{L_{SS}}$ and $\mathcal{L_{WSC}}$ together performs the best. Therefore, although the self-supervision through the controlled transformation forms the backbone of our proposed enhancement module, it draws crucial support from the self-conditioning process where $\mathcal{L_{WSC}}$ plays the dominant role. Ablation 5 clearly depicts that the use of all the three loss functions works the best, and hence, each of the loss functions plays an important role.

\begin{table}[]
\caption{Ablation Study of different loss functions related to our enhancement}
\label{tab: enh_ablation}
\centering
\resizebox{0.45\textwidth}{!}{%
\begin{tabular}{|l|c|c|c|c|}
\hline
Studies &
  $\mathcal{L_{SC}}$ &
  $\mathcal{L_{WSC}}$ &
  $\mathcal{L_{SS}}$ &
  \begin{tabular}[c]{@{}c@{}}PSNR (dB)\end{tabular} 
  \\ \hline
Ablation 1 & \ding{55} & \ding{55} & \ding{51} & 3.77  \\ \hline
Ablation 2 & \ding{51} & \ding{51} & \ding{55} & 9.75  \\ \hline
Ablation 3 & \ding{51} & \ding{55} & \ding{51} & 10.54  \\ \hline
Ablation 4 & \ding{55} & \ding{51} & \ding{51} & 20.27  \\ \hline
\begin{tabular}[c]{@{}l@{}}Ablation 5\\ (SelfEnNet w/o $\mathcal{F_D}$)\end{tabular} &
  \ding{51} &
  \ding{51} &
  \ding{51} &
  20.41  \\ \hline
\end{tabular}%
}
\end{table}

\begin{table}[]
\centering
\caption{Ablation Study of different loss functions related to our noise-handling}
\label{tab: denoise_ablation}
\resizebox{0.4\textwidth}{!}{%
\begin{tabular}{|l|c|c|c|c|}
\hline
Studies & $\mathcal{L_{G}}$ & $\mathcal{L_{F}}$ & $\mathcal{L_{DSC}}$ & \begin{tabular}[c]{@{}c@{}} PSNR (dB)\end{tabular}  \\ \hline
Ablation 6 & \ding{51} & \ding{51} & \ding{55} & 19.97     \\ \hline
Ablation 7 & \ding{51} & \ding{55} & \ding{51} & 21.38  \\ \hline
\begin{tabular}[c]{@{}l@{}}Ablation 8\\ (SelfEnNet)\end{tabular}  & \ding{51} & \ding{51} & \ding{51} & 21.36  \\ \hline
\end{tabular}%
}
\end{table}

\begin{figure}
\vspace{-0.5 cm}
    \centering
    \includegraphics[width=0.35\textwidth]{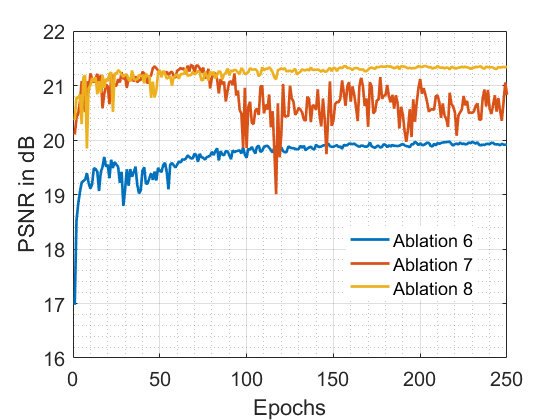}
        \caption{The effect of the different loss functions related to our noise-handling module. The figure shows how PSNR values on test images vary for the trained models after each epoch.}
    \label{fig: ablation_denoise}
\end{figure}

\subsubsection{Importance of the Loss Functions for our Noise-handling Module}
\label{sec:DLoss}
Employing the test images of the LOL dataset, Table~\ref{tab: denoise_ablation} presents the importance of each of the loss functions, namely, low gradient magnitude related loss $\mathcal{L_{G}}$, fidelity loss $\mathcal{L_{F}}$ and denoising self-conditioning loss $\mathcal{L_{DSC}}$ used in our noise-handling module, whose rationale has been provided in Section~\ref{Noise}. The loss $\mathcal{L_{G}}$ is present in all the ablations shown in the table, as without it noise suppression can not take place. The PSNR values shown in the table are obtained after enhancing the denoised image using our enhancement module. Ablations 6 and 7 show the importance of our $\mathcal{L_{DSC}}$ loss function, use of which results in a boost in performance.
From Ablation 8, we see that the result achieved using all the three loss functions is very close to the best obtained by omitting the $\mathcal{L_{F}}$ loss. However, the $\mathcal{L_{F}}$ loss, which is crucial for image detail preservation, ensures that the good performance achieved is not a temporary phenomenon over the number of epochs, which is evident from Fig.~\ref{fig: ablation_denoise}. As given in Table~\ref{tab: denoise_ablation}, Ablation 7 is without $\mathcal{L_{F}}$ loss and Ablation 8 is with $\mathcal{L_{F}}$ loss. We observe from Fig.~\ref{fig: ablation_denoise} that the trained models from the later epochs of Ablation 8 produces consistently higher PSNR values on LOL test images compared to those of Ablation 7.

\subsubsection{Significance of using the Noise-handling Module along with the Enhancement Module}

\begin{figure}
   \centering
   \subfloat[Low-light image]{
     \includegraphics[width=0.14\textwidth]{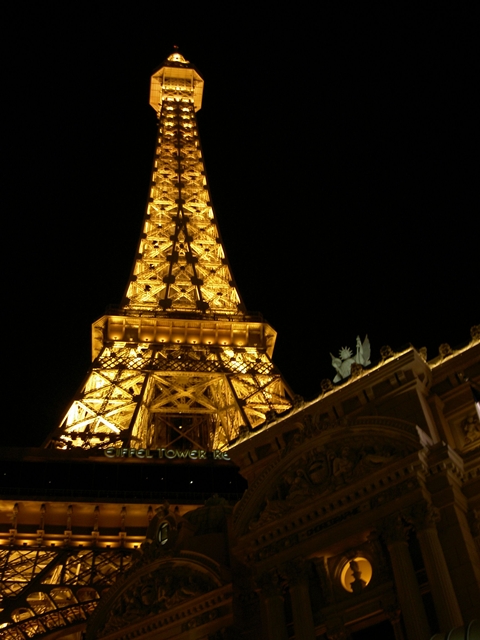}
     \label{fig:Gamma_inp}
   }
   \subfloat[SelfEnNet w/o $\mathcal{F_D}$]{
     \includegraphics[width=0.14\textwidth]{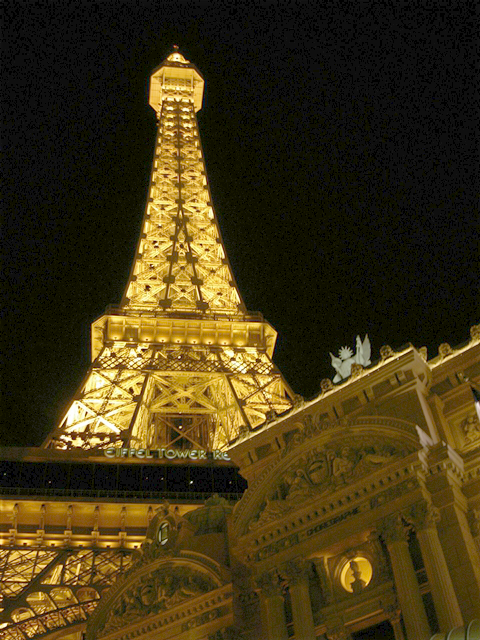}
     \label{fig:Gamma}
   }
   \subfloat[SelfEnNet]{
     \includegraphics[width=0.14\textwidth]{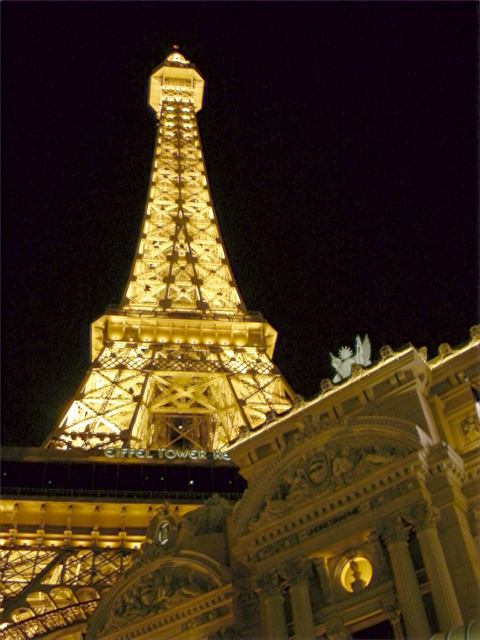}
     \label{fig:Gamma_out}
   }
     \caption{\normalsize Enhancement achieved on (a) a low-light image using SelfEnNet with (b) the enhancement module alone and (c) both the enhancement and noise-handling modules.}
     \label{fig:denoise}
     \vspace{-0.5 cm}
 \end{figure}

Ablation 5 in Table~\ref{tab: enh_ablation} presents the results of our proposed SelfEnNet without the noise-handling module $\mathcal{F_D}$ (only $\mathcal{F_E}$) and Ablation 8 in Table~\ref{tab: denoise_ablation} presents the results of our full SelfEnNet ($\mathcal{F}=\{\mathcal{F_E},\mathcal{F_D}\}$). As can be seen from the two ablations in the two tables, the noise-handling network $\mathcal{F_D}$ helps in gaining about $1$ dB over that achieved by using the enhancement network $\mathcal{F_E}$ alone. In the full network $\mathcal{F}$, $\mathcal{F_E}$ estimates the required degree of enhancement which is followed by $\mathcal{F_D}$ that denoises the input image. Then, the noise-suppressed output from $\mathcal{F_D}$ is enhanced by using the degree of enhancement map from $\mathcal{F_E}$. Fig.~\ref{fig:denoise} depicts the importance of using the noise-handling module along with the enhancement module.

It is evident from the enhanced image in Fig.~\ref{fig:denoise}{\color{red} (c)} that the noise-handling module ($\mathcal{F_D}$) improves upon the result of the significant enhancement achieved by our enhancement module ($\mathcal{F_E}$) shown in Fig.~\ref{fig:denoise}{\color{red} (b)}. The improvement is due to the substantial reduction of noise in the enhanced image, which is especially noticeable comparing the sky regions of the images in Figs.~\ref{fig:denoise}{\color{red} (b)} and {\color{red} (c)}.

\subsubsection{Ablation Studies on the Architecture of Our Approach}
\label{subsec:model_ab}
\paragraph{Number of times of the feedback in the enhancement module}
Table~\ref{tab:feedback} presents an ablation study on the number of times the features from the output is fed back in the enhancement module $\mathcal{F_E}$ (See section~\ref{DSelfLoss}). We vary the number of times ($1-4$) the feedback is given after the initial stage and observe the PSNR values obtained by our SelfEnNet for the low-light test images from the LOL dataset~\cite{yang2021sparse}. The PSNR values shown in the table are computed for the results provided by the models trained with $50$ training epochs. It is evident from the table that we get the best performance with just $1$-time feedback after the initial stage. Hence, we use the same in our approach throughout.

\paragraph{The controlled transformation parameter $\alpha$}
Table~\ref{tab:self_enc} presents an ablation study on the parameter $\alpha$ in the enhancement module. We perform experiments with different values of $\alpha <1$ and again observe the PSNR values obtained for the low-light test images from the LOL dataset~\cite{yang2021sparse}. The  PSNR values shown in the table are for the trained models taken from each of the last $50$ training epochs. with the standard deviations quantifying the corresponding variations. It is evident from the table that the top two PSNR values correspond to $\alpha =0.75$ and $0.85$, respectively. However, we also observed that the standard deviation of the calculated PSNR values using the trained models of the last $50$ epochs for $\alpha=$ $0.75$ and $0.85$ are $0.2384$ and $0.083$, respectively. As the PSNR $20.41\pm0.083$ for $\alpha = 0.75$ indicates a better convergence compared to the PSNR $20.42\pm0.2384$ for $\alpha = 0.85$, we choose $\alpha =0.75$ in our SelfEnNet model.

\begin{table}
\centering
\caption{Performance of the enhancement module with respect to variation in the number of times of the feedback.}
\label{tab:feedback}
\resizebox{0.35\textwidth}{!}{%
\begin{tabular}{|c|c|c|c|c|}
\hline
\begin{tabular}[c]{@{}c@{}}Feedback times\end{tabular}     & 1     & 2     & 3     & 4     \\ \hline
\begin{tabular}[c]{@{}c@{}c@{}}PSNR \end{tabular}          &  \begin{tabular}[c]{@{}c@{}c@{}}20.41 \end{tabular}& \begin{tabular}[c]{@{}c@{}c@{}}20.16\end{tabular}& \begin{tabular}[c]{@{}c@{}c@{}}20.31 \end{tabular}& \begin{tabular}[c]{@{}c@{}c@{}}20.20 \end{tabular}\\ \hline
\end{tabular}%
}
\end{table}

\begin{table}
\caption{An ablation study on the controlled transformation parameter $\alpha$}
\label{tab:self_enc}
\centering
\resizebox{0.45\textwidth}{!}{%
\begin{tabular}{|c|c|c|c|c|c|c|c|}
\hline
\begin{tabular}[c]{@{}c@{}}$\alpha$\end{tabular} & 0.9     & 0.85     & 0.8     & 0.75     & 0.7     & 0.65     & 0.6     \\ \hline
\begin{tabular}[c]{@{}c@{}c@{}}PSNR \end{tabular}          & \begin{tabular}[c]{@{}c@{}c@{}}20.01 \end{tabular} & \begin{tabular}[c]{@{}c@{}c@{}}20.42\end{tabular} & \begin{tabular}[c]{@{}c@{}c@{}}20.34\end{tabular} & \begin{tabular}[c]{@{}c@{}c@{}}20.41\end{tabular} & \begin{tabular}[c]{@{}c@{}c@{}}20.19\end{tabular} & \begin{tabular}[c]{@{}c@{}c@{}}19.97\end{tabular} & \begin{tabular}[c]{@{}c@{}c@{}}19.93 \end{tabular} \\ \hline
\end{tabular}%
}
\end{table}

\begin{table}[h]
\caption{Performance of the enhancement module with respect to different feedback mechanisms \& no feedback.}
\label{tab:model_ablation}
\centering
\resizebox{0.48\textwidth}{!}{%
\begin{tabular}{|c|c|c|c|}
\hline
 &
  \begin{tabular}[c]{@{}c@{}}Feedback \\concatenation\\ with input\end{tabular} &
  \begin{tabular}[c]{@{}c@{}}Feedback\\ concatenation\\ with features\end{tabular} &
  \begin{tabular}[c]{@{}c@{}}PSNR\\ in dB\end{tabular} \\ \hline
Ablation 1         & \ding{55} & \ding{55} & 13.37 \\ \hline
Ablation 2          & \ding{51} & \ding{55} & 20.06 \\ \hline
Ablation 3 ($\mathcal{F_E}$)  & \ding{55} & \ding{51} & 20.41 \\ \hline
\end{tabular}%
}
\end{table}

\color{black}
\paragraph{The feedback type in the enhancement module}
In the architecture of our enhancement module, the feedback features from the output are concatenated with those from the first input convolutional layer, whose benefit is mentioned in Section~\ref{DSelfLoss}. This is preferred over the direct concatenation of the output with the input image to the module. Table~\ref{tab:model_ablation} shows the ablation study on the architecture of the enhancement module $\mathcal{F_E}$ that led us to the said choice of feedback mechanism. Ablation 1 in the table represents the model of the module where no feedback mechanism is used. Ablation 2 represents the model which feedbacks its output as it is to the input, where they are concatenated before any feature extraction (by input convolutional layers). We can observe that feedback plays a crucial role as Ablation 1 performs poorly in comparison to Ablation 2. Ablation 3 represents the model which feedbacks features extracted from its output to concatenate with the features from the first input convolutional layer. As can be seen from the table, this concatenation of the features results in the best performance compared to the other two cases. Hence, we choose the model represented by Ablation 3 for our enhancement module.

\begin{table}
\caption{Performance of the noise-handling module with respect to different ways of using the enhancement map $\eta_I$.}
\label{tab:denoise_model_ablation}
\centering
\resizebox{0.5\textwidth}{!}{%
\begin{tabular}{|c|c|c|c|}
\hline
 &
  \begin{tabular}[c]{@{}c@{}}$\eta_I$ concatenation\\ with input\end{tabular} &
  \begin{tabular}[c]{@{}c@{}}$\eta_I$ concatenation\\ with features\end{tabular} &
  \begin{tabular}[c]{@{}c@{}}PSNR\\ in dB\end{tabular} \\ \hline
Ablation 1         & \ding{55} & \ding{55} & 21.22 \\ \hline
Ablation 2         & \ding{51} & \ding{55} & 21.28 \\ \hline
Ablation 3 ($\mathfrak{F_D}$)        & \ding{55} & \ding{51} & 21.36 \\ \hline
\end{tabular}%
}
\end{table}

\paragraph{The use of enhancement map in the noise-handling module}
\label{ablation_enh_map}
In the model of our noise-handling module $F_D$, the enhancement map estimated from the input low-light image by the trained enhancement module is used, whose benefit is mentioned in Sections~\ref{Noise} and~\ref{enh_map_feat}. In the model architecture, features extracted from the enhancement map are concatenated with the features from the low-light image obtained using first input convolutional layer. This is preferred over directly concatenating the enhancement map with the input image to the module. Table~\ref{tab:denoise_model_ablation} shows the ablation study of the noise-handling module's architecture that guided us towards the said choice. Ablation 1 in the table represents the model where the enhancement map is not used. Ablation 2 represents the model where the map is concatenated with the input image before any feature extraction. Ablation 3 represents the model where the features extracted from the map and the input image are concatenated. As we observe from the table, Ablation 3 performs slightly better than the other two cases, suggesting the usefulness of using the enhancement map for denoising when used as represented by Ablation 3. Hence, we choose the model represented by Ablation 3 for our noise-handling module.

\begin{table}[!t]
\caption{\color{black}Performance of our approach in the LOL test images when trained on low-light images from different datasets}
\label{tab: dataset}
\centering
\resizebox{0.5\textwidth}{!}{
\begin{tabular}{|c|c|c|c|}
\hline
  {\begin{tabular}[c]{@{}c@{}}Training $\rightarrow$\\ Measure $\downarrow$ \end{tabular}} &
  \multicolumn{1}{c|}{\begin{tabular}[c]{@{}c@{}}Low-light images: ZeroDCE\\ Well-lit images: LOL\end{tabular}} &
  \multicolumn{1}{c|}{\begin{tabular}[c]{@{}c@{}}Low-light images: RLL\\ Well-lit images: LOL\end{tabular}} & \multicolumn{1}{c|}{\begin{tabular}[c]{@{}c@{}}Low-light images: LOL\\ Well-lit images: LOL\end{tabular}} \\ \hline
PSNR  & 21.04 & 20.98 & 21.36 \\ \hline
SSIM  & 0.76  & 0.76 & 0.77 \\ \hline
CIEDE & 39.33 & 40.15 & 39.43  \\ \hline
LPIPS$_\textrm{VGG}$ & 0.33  & 0.33 & 0.31   \\ \hline
\end{tabular}}
\end{table}

\begin{table*}[]
\caption{Performance comparison on the $500$ real-world test images from MIT-Adobe 5K dataset~\cite{fivek}. {\color{red}\textbf{Red}} Highlight: Best, {\color{black}\textbf{Blue}} Highlight: Second Best.}
\label{tab: MIT}
\resizebox{\textwidth}{!}{
\begin{tabular}{|l|lllll|llllllllll|}
\hline
\multirow{3}{*}{Measures} &
  \multicolumn{5}{c|}{\multirow{2}{*}{Paired Supervision}} &
  \multicolumn{10}{c|}{No paired Supervision} \\ \cline{7-16} 
 &
  \multicolumn{5}{c|}{} &
  \multicolumn{8}{c|}{No Unpaired Supervision} &
  \multicolumn{2}{l|}{Unpaired Supervision} \\ \cline{2-16} 
 &
 \multicolumn{1}{l|}{\begin{tabular}[c]{@{}l@{}}DRD~\cite{wei2018deep}\\ BMVC'21\end{tabular}} & \multicolumn{1}{l|}{\begin{tabular}[c]{@{}l@{}}DRBN~\cite{yang2021band}\\ TIP'21\end{tabular}}&
\multicolumn{1}{l|}{\begin{tabular}[c]{@{}l@{}}URN~\cite{wu2022uretinex}\\ CVPR'22\end{tabular}}&
\multicolumn{1}{l|}{\begin{tabular}[c]{@{}l@{}}MLLEN~\cite{fan2022multiscale}\\ TCSVT'22\end{tabular}}&
  \begin{tabular}[c]{@{}l@{}}Bread~\cite{guo2023low}\\ IJCV'23\end{tabular}  &
  \multicolumn{1}{l|}{\begin{tabular}[c]{@{}l@{}}LIME~\cite{Guo2017}\\ TIP'17\end{tabular}} &
  \multicolumn{1}{l|}{\begin{tabular}[c]{@{}l@{}}RRM~\cite{li2018structure}\\ TIP'18\end{tabular}} &
  \multicolumn{1}{l|}{\begin{tabular}[c]{@{}l@{}}ALSM~\cite{wang2019low}\\ TIP'19\end{tabular}} &
  \multicolumn{1}{l|}{\begin{tabular}[c]{@{}l@{}}Zero-DCE~\cite{guo2020zero}\\ CVPR'20\end{tabular}} &
  \multicolumn{1}{l|}{\begin{tabular}[c]{@{}l@{}}ZCP$^*$~\cite{kar2021zero}\\ CVPR'21\end{tabular}} &
  \multicolumn{1}{l|}{\begin{tabular}[c]{@{}l@{}}RUAS~\cite{liu2021retinex}\\ CVPR'21\end{tabular}} &
  \multicolumn{1}{l|}{\begin{tabular}[c]{@{}l@{}}SCI~\cite{ma2022toward}\\ CVPR'22\end{tabular}} &
  \multicolumn{1}{l|}{\begin{tabular}[c]{@{}l@{}}PairLIE~\cite{fu2023learning}\\ CVPR'23\end{tabular}} &
  \multicolumn{1}{l|}{\begin{tabular}[c]{@{}l@{}}EGAN~\cite{jiang2021enlightengan}\\ TIP'21\end{tabular}} &
  \multicolumn{1}{l|}{\begin{tabular}[c]{@{}l@{}}SelfEnNet\\ (Ours)\end{tabular}} \\ \hline
  \begin{tabular}[c]{@{}l@{}}PSNR\\ SSIM\\ CIEDE\\ LPIPS$_\textrm{VGG}$\\ NIQE\\ LOE\end{tabular} &
  \multicolumn{1}{l|}{\begin{tabular}[c]{@{}l@{}}12.41 \\0.71 \\38.85 \\0.23 \\1717.25 \\4.12 \end{tabular}} & 
  \multicolumn{1}{l|}{\begin{tabular}[c]{@{}l@{}}16.55 \\0.74 \\45.28 \\0.25 \\721.17 \\3.28 \end{tabular}} & 
  \multicolumn{1}{l|}{\begin{tabular}[c]{@{}l@{}}13.79 \\0.71 \\43.75 \\0.21 \\675.18 \\3.20 \end{tabular}} & 
  \multicolumn{1}{l|}{\begin{tabular}[c]{@{}l@{}}15.56 \\0.53 \\54.58 \\0.35 \\547.11 \\3.14 \end{tabular}} & 
  \begin{tabular}[c]{@{}l@{}}17.06 \\0.74 \\35.68 \\0.24 \\555.91 \\3.56 \end{tabular} & 
  \multicolumn{1}{l|}{\begin{tabular}[c]{@{}l@{}}17.02 \\{\color{red}\textbf{0.81}} \\{\color{red}\textbf{31.34}} \\{\color{red}\textbf{0.12}} \\574.45 \\3.17 \end{tabular}} & 
  \multicolumn{1}{l|}{\begin{tabular}[c]{@{}l@{}} {\color{red}\textbf{17.68}} \\{\color{black}\textbf{0.79}} \\{\color{black}\textbf{31.66}} \\0.19 \\444.51 \\3.62 \end{tabular}} & 
  \multicolumn{1}{l|}{\begin{tabular}[c]{@{}l@{}}15.36 \\0.75 \\40.07 \\{\color{black}\textbf{0.13}} \\658.11 \\3.15 \end{tabular}} & 
  \multicolumn{1}{l|}{\begin{tabular}[c]{@{}l@{}}15.78 \\0.69 \\47.49 \\0.17 \\444.77 \\3.09 \end{tabular}} & 
  \multicolumn{1}{l|}{\begin{tabular}[c]{@{}l@{}}13.73 \\0.62 \\38.87 \\0.26 \\1135.05 \\4.07 \end{tabular}} & 
  \multicolumn{1}{l|}{\begin{tabular}[c]{@{}l@{}}7.52 \\0.39 \\61.01 \\0.47 \\1459.24 \\5.13 \end{tabular}} & 
  \multicolumn{1}{l|}{\begin{tabular}[c]{@{}l@{}}12.64 \\0.68 \\46.93 \\0.15 \\{\color{black}\textbf{145.70}} \\3.07 \end{tabular}} & 
  \multicolumn{1}{l|}{\begin{tabular}[c]{@{}l@{}}13.18 \\0.70 \\47.05 \\0.24 \\359.97 \\4.00 \end{tabular}} & 
  \multicolumn{1}{l|}{\begin{tabular}[c]{@{}l@{}}15.48 \\0.74 \\45.43 \\0.16 \\755.55 \\{\color{black}\textbf{3.06}} \end{tabular}} & 
 {\begin{tabular}[c]{@{}l@{}} {\color{black}\textbf{17.19}} \\0.68 \\37.87 \\0.20 \\{\color{red}\textbf{111.38}} \\{\color{red}\textbf{2.94}} \end{tabular}} \\ \hline 
\end{tabular}}
\vspace{-0.3 cm}
\end{table*}

\subsection{Additional Experimental Studies}
\label{crossData}

In Section~\ref{subsec:result}, it is shown that as our approach is trained using low-light and well-lit images in an unpaired manner, the mapping from low-light to well-lit images learned by it is robust unlike that learned by approaches trained through paired supervision, and this results in its better performance in a wide variety of low-light images from different datasets. 
 
Now, as our approach is trained in an unpaired manner, it is expected to perform similarly even when trained with different kinds of low-light images that are not related in any manner to the kind of well-lit images desired. Table~\ref{tab: dataset} depicts results from an experiment that tests our approach while training it on low-light images from different datasets and on well-lit images of the desired kind in an unpaired manner. As can be seen, low-light images from `ZeroDCE' dataset of \cite{guo2020zero} and `RLL' dataset of \cite{RLLdataset} are used apart from that in the LOL training image set, and well-lit images of the LOL training set are employed considering them as the desired kind. The quantitative measures shown are obtained by applying the differently trained models on the low-light images of the LOL testing set. It is evident that there is only a marginal drop in the performance when the low-light images from the `ZeroDCE' or `RLL' datasets are used instead of those from the LOL dataset for the training, as it was expected earlier. Note that this observation also signifies the robustness of the low-light image to well-lit image mapping learned by our approach like the observation from Table~\ref{tab: Real} earlier.

\color{black}
We perform another experiment in order to compare the robustness of our method to that of the others. We use  MIT-Adobe 5K dataset~\cite{fivek} as a test set where none of the methods are trained.  As the images of this dataset are tone-mapped using ``software dedicated to photo adjustment (Adobe Lightroom)"~\cite{fivek}, the dataset image distribution is different from that in the training dataset. Table~\ref{tab: MIT} shows the performance of our method as compared to the other modern techniques. Deep learning based methods are prone to fail in generalizing on datasets that are out of training distribution \cite{zhang2021understanding}. This seems to be the case in the table where the sophisticated hand-crafted models LIME and RRM produce the better results among the existing methods. However, our deep learning based method does as good as or better than these hand-crafted models in Table~\ref{tab: MIT}, while outperforming them convincingly in $6$ low-light image enhancement datasets as evident from Tables~\ref{tab: LOL} and~\ref{tab: Real}. These observations together eventually indicate the robustness of our approach and its superiority.

\color{black}

\subsection{Limitation}
 In our work, the learning of the enhancement module via self-conditioning is performed by attempting not to enhance well-lit images as they do not require enhancement, and by avoiding further enhancement of low-light images that have been already enhanced using the model. Our aim here is to enhance low-light images through unpaired supervision based only on low-light image enhancement-specific attributes over any generic perceptual regularizer. Due to this, our enhanced images  sometimes may look a bit visually less vibrant, which may be considered a limitation for some applications. The goal of our approach is of sufficient and consistent enhancement rather than to produce vibrant, sometimes over-enhanced, images. The above can be observed in the last few rows of Fig.~\ref{fig:SubjGT}, where although our technique produces enhanced images closer to the corresponding ground truths, they are not as vibrant as the images produced by a few other techniques. 

\color{black}
\section{Conclusion}
\label{sec:conclude}
This paper contributes a low-light image enhancement approach using novel self-supervision and self-conditioning strategies. The proposed deep network, which has a noise-handling module apart from the main enhancement module, is trained using unpaired low-light and well-lit images based on loss functions specific to the proposed strategies. While the self-supervision process attempts to achieve consistency in enhancement, the self-conditioning process aims for the sufficiency of it. Detailed studies show the effectiveness of both the strategies, especially the self-supervision through the controlled transformation that is the backbone of our proposed enhancement network. Our approach is found to perform satisfactory enhancement without introducing any visible distortion. It is also found to be good at avoiding over-enhancement and in producing natural-looking detail-preserving enhanced images. The success of our controlled transformation based unpaired self-conditioning and self-supervision strategies pave the way for further studies on such effective ground truth independent image restoration approaches.

\section{Aknowledgement}
Debashis Sen acknowledges the Science and Engineering Research Board (SERB), India for its assistance.

\ifCLASSOPTIONcaptionsoff
  \newpage
\fi


\end{document}